\documentclass[sigconf, nonacm]{acmart}

\renewcommand\footnotetextcopyrightpermission[1]

\usepackage{tcolorbox}
\usepackage{xspace}
\newcommand\vldbdoi{XX.XX/XXX.XX}

\newcommand{\model}{\texttt{Fremer}\xspace}

\newcommand\vldbavailabilityurl{https://github.com/YHYHYHYHYHY/Fremer}
 
\usepackage{multirow}
\usepackage{threeparttable} 
\usepackage{hyperref}

\begin{document}

\title{Fremer: Lightweight and Effective Frequency Transformer for Workload Forecasting in Cloud Services}

\author{
    Jiadong Chen$^{1,3,*}$,
    Hengyu Ye$^{1,*}$,
    Fuxin Jiang$^{2}$,
    Xiao He$^{2}$,\\ 
    Tieying Zhang$^{2}$,
    Jianjun Chen$^{2}$,   
    Xiaofeng Gao$^{1, \dagger}$
}
\affiliation{
  $^1$ Shanghai Jiao Tong Univerisity;
  $^2$ ByteDance Inc.;
  $^3$ University of New South Wales;
}

\email{
  {cs_22_yhy,chenjiadong998}@sjtu.edu.cn,  gao-xf@cs.sjtu.edu.cn
}
\email{
  {jiangfuxin, xiao.hx, tieying.zhang, jianjun.chen}@bytedance.com
}

\begin{abstract}
Workload forecasting is pivotal in cloud service applications, such as auto-scaling and scheduling, with profound implications for operational efficiency. 
Although Transformer-based forecasting models have demonstrated remarkable success in general tasks, their computational efficiency often falls short of the stringent requirements in large-scale cloud environments. Given that most workload series exhibit complicated periodic patterns, addressing these challenges in the frequency domain offers substantial advantages. To this end, we propose \model, an efficient and effective deep forecasting model. \model\ fulfills three critical requirements: it demonstrates superior efficiency, outperforming most Transformer-based forecasting models; it achieves exceptional accuracy, surpassing all state-of-the-art (SOTA) models in workload forecasting; and it exhibits robust performance for multi-period series. Furthermore, we collect and open-source four high-quality, open-source workload datasets derived from ByteDance's cloud services, encompassing workload data from thousands of computing instances. Extensive experiments on both our proprietary datasets and public benchmarks demonstrate that \model\ consistently outperforms baseline models, achieving average improvements of 5.5\% in MSE, 4.7\% in MAE, and 8.6\% in SMAPE over SOTA models, while simultaneously reducing parameter scale and computational costs. Additionally, in a proactive auto-scaling test based on Kubernetes, \model\ improves average latency by 18.78\% and reduces resource consumption by 2.35\%, underscoring its practical efficacy in real-world applications.
\end{abstract}

\maketitle

\renewcommand\thefootnote{}\footnote{\noindent
* indicates equal contribution. $\dagger$ Xiaofeng Gao is the corresponding author.}


\ifdefempty{\vldbavailabilityurl}{}{
\vspace{.3cm}
\begingroup\small\noindent\raggedright\textbf{PVLDB Artifact Availability:}\\
{The source code has been made available at \url{\vldbavailabilityurl}. The ByteDance Cloud workload datasets have been made available at {\hyperlink{https://huggingface.co/datasets/ByteDance/CloudTimeSeriesData}{https://huggingface.co/datasets/ByteDance/CloudTimeSeriesData}}.}
\endgroup
}
\setcounter{section}{0}
\setcounter{figure}{0}
\setcounter{table}{0}
\setcounter{equation}{0}
\section{Introduction}\label{sec:intro}

\begin{figure}[h]
\centering
\includegraphics[width=0.5\textwidth]{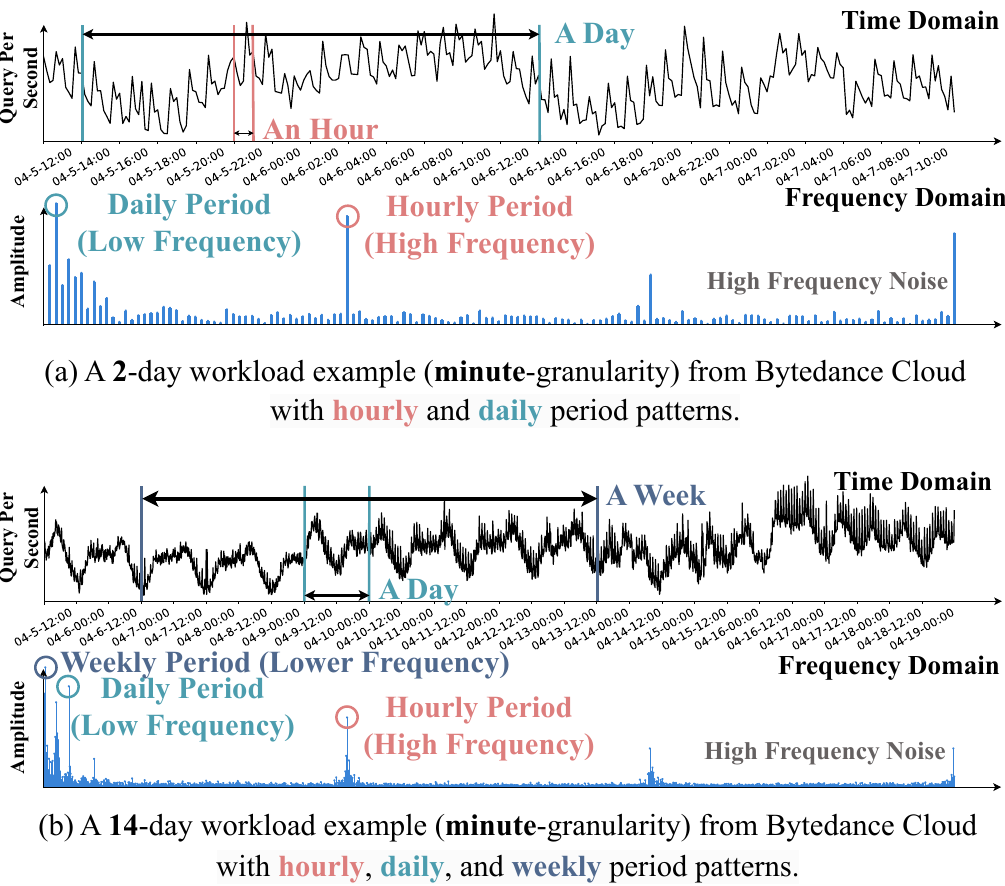}
\centering
\caption{{Workload from cloud services in ByteDance exhibit complicated periodicity, both in time and frequency domain.}}
\label{fig:intro}
\vspace{-2pt}
\end{figure}
In modern cloud service platforms, tens of thousands of applications and millions of microservices are deployed to support diverse workloads. To ensure service quality (QoS) and optimize resource utilization, these platforms rely on monitoring systems that collect and analyze workload patterns. Accurate workload forecasting is critical for enabling predictive active scaling technologies, which are widely adopted to meet dynamic demand~\citep{moore2013transforming,lombardi2019pascal,golshani2021proactive,shariffdeen2016adaptive,lamas2016purely,alahmad2021proactive}.

However, achieving both efficiency and effectiveness in workload forecasting poses significant challenges in large-scale cloud environments. For instance, the Platform-as-a-Service (PaaS) of Bytedance Cloud requires the execution of over 100,000 forecasting tasks per hour, demanding efficient models that can operate with minimal computational overhead. 
{Figure 1 illustrates workload series from a real-world cloud computing system, showing minute-granularity patterns over 2 days (Figure 1(a)) and 14 days (Figure 1(b)). The cloud system supports diverse services like web applications, real-time data processing, and machine learning tasks, thus inherently handling highly complex workloads series in the time domain with hourly, daily, and weekly temporal patterns. In the frequency domain, different periodic components and high-frequency noise are more easily distinguishable. Moreover, time series data with significant variation and hard-to-quantify similarity in the time domain appears more similar in the frequency domain. Frequency-domain information is often concentrated near a few frequencies and exists in combinations, such as harmonics, as shown in the lower parts of Figure 1(a) and 1(b). These characteristics motivate our use of frequency-domain methods for fine-grained, large-scale workload series forecasting.}

\begin{figure}[h]
\centering
\includegraphics[width=0.48\textwidth]{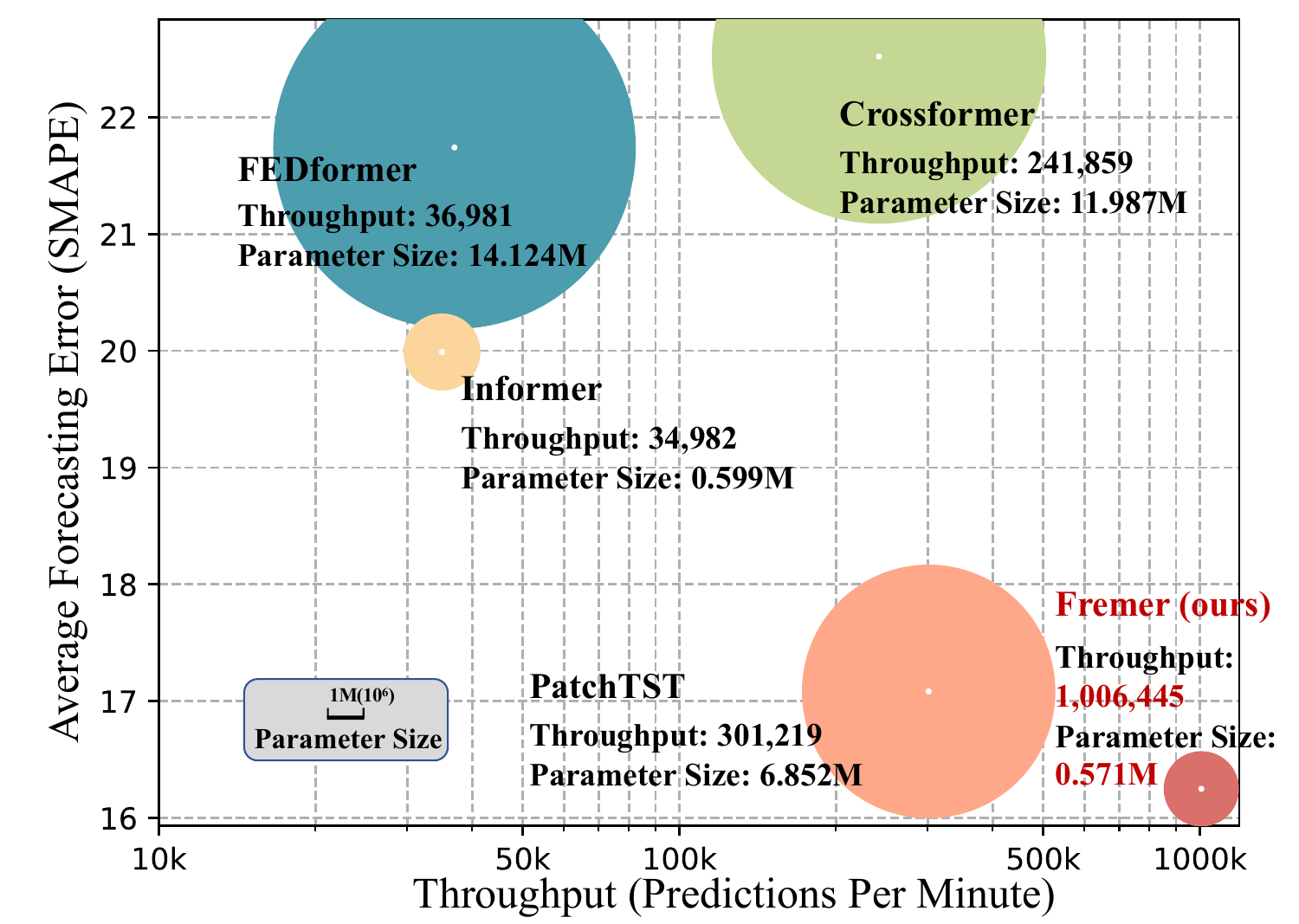}
\centering
\caption{Comparison of throughput and prediction error among different Transformers on IaaS workload dataset (experimental settings are referred to in Section~\ref{sec:expsetting}).}
\label{fig:intro_throu}
\vspace{-0pt}
\end{figure}

Among existing time-series forecasting models, those based on the Transformer architecture have achieved state-of-the-art (SOTA) performance. This is largely attributed to the attention mechanism, which enables effective modeling of long sequences~\citep{zhou2021informer, wu2021autoformer,zhang2022crossformer, nie2022time}. However, these methods often struggle to capture multi-periodic patterns in complex time series, particularly when fine-grained periodicity (e.g., minute-level) intertwines with broader cycles (e.g., daily or weekly), leading to suboptimal forecasting accuracy, as shown in Figure~\ref{fig:intro_throu}. Meanwhile, the efficiency drawbacks of Transformer limit its deployment in real-world scenarios.


The complexity of attention mechanism is quadratically proportional to the length of the input series, which restricts their deployment in dealing with long sequence predictions. Moreover, the inherent susceptibility of neural networks to overfitting poses significant challenges in achieving robust generalization~\citep{zeng2023transformers}, particularly when handling large-scale and complex time series data. Figure~\ref{fig:intro_throu} illustrates the throughput and accuracy of different Transformer-based models, as well as their parameter volume. Existing works either have low accuracy or are affected by parameter volume or computational efficiency that affect total throughput.

The efficiency bottleneck of Transformer-based methods lies primarily in the computation of attention. For a sequence of length $L$, the time and space complexity of the attention mechanism are both $O(L^2)$, resulting in significant inference time and GPU memory usage. Furthermore, beyond the model, the time domain representation of series is also a barrier to efficient and accurate data mining. Research indicates that obtaining a compact representation of sequences in the time domain is relatively difficult~\citep{zhou2022fedformer}. In contrast, sequence data can achieve a compact representation in the frequency domain~\citep{zhou2022film}, as noise, trends, and periodic information are effectively decoupled in this domain. Moreover, the frequency domain allows for better capture of global information (such as period shift)~\citep{zhou2022fedformer}, as demonstrated in Figure~\ref{fig:intro}.



Transforming time series into the frequency domain can provide a foundation for efficient mining of sequence information, but there are still technical challenges that need to be addressed. 
\textbf{1)~The issue of {frequency mis-alignment}}. 
The Discrete Fourier Transform (DFT) is a common method for transforming time series from the time domain to the frequency domain.
Since the frequency sampling interval of the DFT is fixed (for a sequence of length \( L \), the sampling interval is \( 1/L \)), the true periodic frequency of the sequence may not be sampled, as shown in Figure~\ref{fig:FreqMis}. However, this issue has not yet received sufficient attention from the community. To the best of our knowledge, we are the first to address this problem.
\textbf{2)~The challenge of extracting key frequency information}. Information in the frequency domain often exists in a combination form (e.g., harmonics). The global information of a sequence often originates from the combination of a few key periodic features, especially when the sequence contains multiple periodic patterns as shown in Figure~\ref{fig:intro}. Recent works~\citep{zhou2022fedformer, lin2024sparsetsf} model point-to-point relationships between frequency points while neglecting the issue of frequency-domain information appearing as combinations. Therefore, it is necessary to explore the relationships between frequency combinations.

In this paper, we propose \model, an efficient and effective forecasting model.
\model\ aims at forecasting the spectrum of the complete series (including input part and forecasting part) based on spectrum of input series.
Specifically, targeting at the unique characteristics of frequency domain representation, we design \textbf{1)~Learnable Linear Padding} for addressing the frequency resolution mis-alignment problem; \textbf{2)~Complex-valued Spectrum Attention} to effectively capture global dependencies from the frequency combinations; \textbf{3)~Frequency Filters} for handling noise and overfitting.




We also provide and open-source four high-quality workload datasets\footnote{\hyperlink{https://huggingface.co/datasets/ByteDance/CloudTimeSeriesData}{https://huggingface.co/datasets/ByteDance/CloudTimeSeriesData}} from distinct cloud service types in ByteDance, containing workload data of thousands of computing instances, spanning from 1 month to 2 months. We carefully preprocess the raw data, and make them useful tools for the community to evaluate and develop new workload forecasting methods. Through extensive experiments on these datasets and three public datasets, our proposed \model\ achieved an average improvement of 5.5\% in MSE, 4.7\% in MAE and 8.6\% in SMAPE compared to current SOTA forecasting models, with scale of parameters and computing costs reduced by times. 

Our contributions are summarized as follows:
{\begin{itemize}
    \item We proposed \model, a deep forecasting framework for workload forecasting, effectively utilizing the frequency domain representation to achieve the balance among accuracy, efficiency, and generalizability.
    \item We design Learnable Linear Padding, Frequency Filters, and Complex-valued Spectrum Attention, with which \model\ outperforms all SOTA forecasting models.
    \item {We open-source four workload datasets from ByteDance’s cloud services, containing workload data from thousands of computing instances over 1–2 months, providing robust support for training and evaluating forecasting models.}
    \item Extensive experiments on the Time Series Forecasting Benchmark (TFB)\citep{qiu2024tfb} and our datasets demonstrate \model’s superiority. 
    It achieves performance improvements across datasets in various domains while significantly reducing parameter scale and computational costs. In Kubernetes HPA proactive auto-scaling tests, \model\ reduces average latency by 18.78\% and resource consumption by 2.35\%, validating its real-world efficacy.
\end{itemize}}

\section{Related Work}\label{sec:related}

\textbf{Overview of Forecasting Methods.}
Early time series forecasting predominantly relied on traditional statistical methods, such as ETS~\citep{holt2004forecasting}, ARIMA~\citep{box1968some,kumar2016forecasting, calheiros2014workload,li2013workload,khorsand2018withdrawn}, STL~\citep{DBLP:journals/pvldb/HeLTWL23}, and regression-based methods~\citep{antonescu2016simulation, yang2013workload,barati2015hybrid,raghunath2015virtual}. However, these methods often have limitations in capturing complex temporal dynamics.
In recent years, deep learning has significantly propelled the field of time series forecasting~\citep{nguyen2016workload,khan2022workload,saxena2021workload, chung2014empirical}. RNN-based methods \citep{lai2018modeling, song2018host,ruta2020deep,leka2021hybrid,karim2021bhyprec} and CNN-based methods \citep{bai2018empirical, liu2022scinet} have shown enhanced abilities in extracting intricate patterns from time series data.
Additionally, MLP-based methods \citep{oreshkin2019n, challu2022nhits, zeng2023transformers} offer a straightforward yet effective approach for learning the non-linear temporal dependencies. {The frequency domain offers a unique perspective for analyzing time series data by revealing underlying periodicity and frequency components. Several forecasting methods have been developed to exploit these characteristics and capture global dependencies inherent in time series data, such as FECAM~\citep{jiang2023fecam}, FilM~\cite{zhou2022film}, FreDo~\citep{sun2022fredo}, FreTS~\citep{yi2023frequency}, FilterNet~\citep{yi2024filternet} and FITS~\citep{xu2024fits}.}

{\textbf{Transformer-based Forecasting Methods.}
Transformers have achieved breakthroughs in many tasks, and attention mechanisms have outperformed traditional time series analysis methods. Based on model architecture and design concepts, we categorize Transformer-based forecasting methods into four groups: long-sequence modeling, sequence decomposition, spatial dependency, and periodicity-aware approaches. Long-sequence modeling methods (Informer~\cite{zhou2021informer}, PatchTST~\cite{nie2022time}, Reformer~\cite{kitaev2020reformer}, Pyraformer~\cite{liu2022pyraformer}) efficiently capture long-range temporal dependencies. Sequence decomposition methods (FEDformer~\cite{zhou2022fedformer}, Autoformer~\cite{wu2021autoformer}, ETSformer~\cite{woo2022etsformer}, Basisformer~\cite{ni2023basisformer}) decompose sequences into trend, periodic, and residual components for higher forecasting accuracy. Spatial dependency methods (Crossformer~\cite{zhang2022crossformer}, iTransformer~\cite{liu2023itransformer}, Earthformer~\cite{gao2022earthformer}, Airformer~\cite{liang2023airformer}) model variable correlations using spatio-temporal information. Frequency and periodicity-aware methods (Fredformer~\cite{piao2024fredformer}, PDF~\cite{dai2024periodicity}) extract periodicity information via Discrete Fourier Transform to explore intra-period patterns and inter-period relationships.}

\section{Preliminary}


\subsection{Workload Forecasting}
\textbf{Definition of Workload Series}
In the context of cloud computing, a workload series represents a time-ordered aggregation of workloads generated by jobs or applications operating on cloud infrastructure~\citep{alahmad2021proactive}.
The Common types of workload in cloud environment, including distinct resource-consumption patterns (such as CPU and memory usage) and user-request metrics (like Queries Per Second-QPS). For a computing instance, the workload series can be mathematically defined as $\mathbf{X}=\{x_0,x_1,\dots,x_{L-1}\}$, where $L$ denotes the length of a given time period, which is equivalent to the number of time steps utilized as input, and $w_i$ represents the numerical value of the specific workload at time-step $i$.

\textbf{Workload Forecasting.} Given the historical value of a workload series $\mathbf{X}=\{x_0,x_1,\dots,x_{L-1}\}$,  the objective of workload forecasting is to predict the future workload series $\mathbf{\hat{Y}}=\{x_{L},x_{L+1},\dots,x_{L+T-1}\}$. In this expression, $T$ represents the length of the forecasting horizon, that is, the number of time steps to be predicted.
The target of workload forecasting is to make the prediction $\mathbf{\hat{Y}}$ as accurate as possible, which is equivalent to minimizing the gap between the prediction $\mathbf{\hat{Y}}$ and the corresponding ground truth value $\mathbf{Y}$.

\subsection{Time Series in Frequency Domain}
The frequency domain of a workload series refers to the representation of the data in terms of its frequency components. The frequency domain analysis provides insights into the underlying periodicity and patterns present in the workload series data that may not be readily apparent in the time domain.

\textbf{Discrete Fourier Transform}
The Discrete Fourier Transform (DFT) is a mathematical transformation that converts a time series into its frequency domain representation, revealing the frequency components present in the data~\citep{Sundararajan2025}. 
Specifically, for an input series with length $L$ , the DFT spectrum is calculated as follows:
\begin{equation}
    \mathbf{F}[k]=\sum_{n=0}^{L-1}\mathbf{X}[n]e^{-2\pi i\frac{kn}{L}}, k=0,1,\dots,L-1.
    \label{DFT}
\end{equation}
Here, $\mathbf{F}[k]$ represents the frequency-domain representation (the spectrum) at frequency index $k$, $\mathbf{X}$ represents the time-domain input sequence, $L$ is the number of data points in the sequence, and $i$ is the imaginary unit ($i^2=-1$).

The Inverse Discrete Fourier Transform (iDFT) is the reverse operation of the DFT. It takes a frequency-domain signal, obtained through the DFT, and reconstructs the original time-domain signal:
\begin{equation}
    \mathbf{X}[n]=\frac{1}{L}\sum_{k=0}^{L-1}\mathbf{F}[k]e^{2\pi i\frac{kn}{L}}, n=0,1,\dots,L-1.
\end{equation}

The Fast Fourier Transform (FFT) is an efficient algorithm for computing the DFT and iDFT. In practice, the Real FFT (rFFT) is typically used for transforming real-valued data, as it exploits Hermitian symmetry to compute only the non-redundant positive frequencies, reducing computational cost and memory usage.

\textbf{Frequency Filters.}
Frequency filters selectively enhance or suppress specific frequency components in a signal. A High-Pass Filter (HPF) attenuates low-frequency components, isolating rapid changes or fine details, while a Low-Pass Filter (LPF) attenuates high-frequency components, smoothing noise and capturing broader trends. These filters are indispensable in signal processing, enabling precise control over frequency ranges to meet specific needs.

\section{Proposed Method}



In this section, we propose \model, which re-designs the classical Transformer into an Encoder-only architecture based on frequency representation. This architecture is specialized for workload forecasting from the perspective of the frequency domain. \model\ incorporates two key designs that effectively address the challenges in frequency-domain forecasting: 1) Learnable Linear Padding (LLP) for frequency resolution alignment; 2) Complex-valued Attention (CSA) for extracting the relationships between frequency combinations and reducing complexity. Moreover, \model\ is designed in a channel-independent manner, and its effectiveness has been demonstrated by \citep{nie2022time}. This design also endows \model\ with generalizability, enabling it to be easily applied to workload forecasting for unseen instances.

\begin{figure}[h]
\centering
\includegraphics[width=0.48\textwidth]{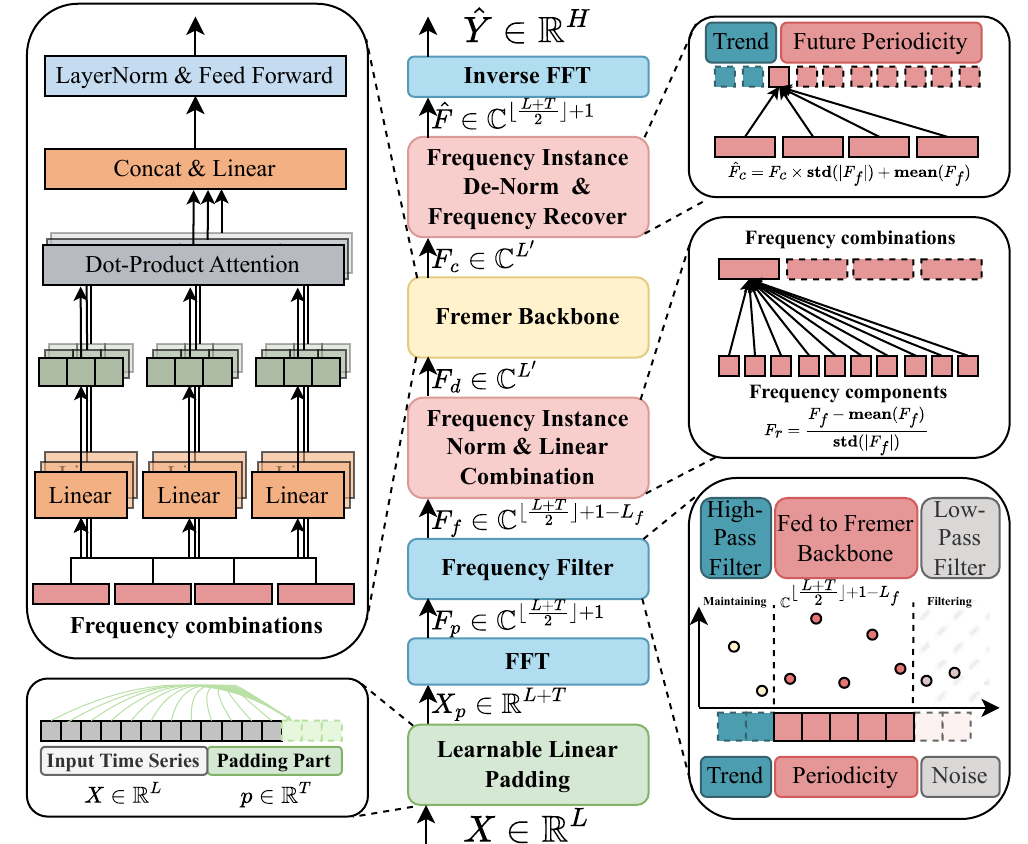}
\centering
\caption{ {Model Architecture of \model.}}
\label{fig:Fremer}
\vspace{-4mm}
\end{figure}


\begin{figure*}[t]
\centering
\includegraphics[width=\textwidth]{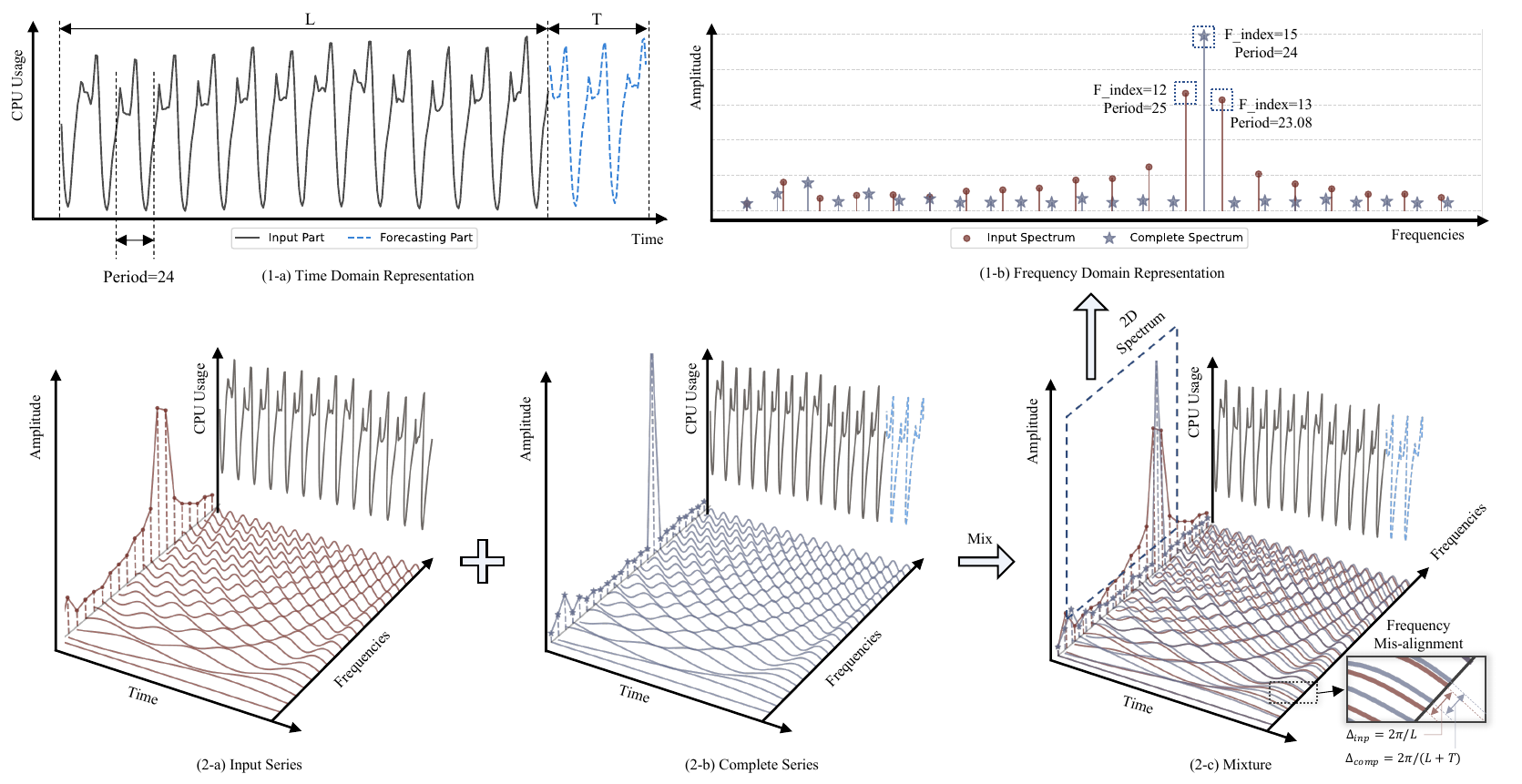}
\centering
\caption{ Frequency Resolution Mis-alignment. The original workload series is sourced from PaaS dataset. It could be observed that the spectrum of input series and complete series mis-align in resolution.}
\label{fig:FreqMis}
\vspace{-0pt}
\end{figure*}
As illustrated in Figure \ref{fig:Fremer}, the entire pipeline of \model\ can be described as follows:
Given an input workload series  $\mathbf{X}\in \mathbb{R}^{L}$, we use LLP to pad it to $\mathbf{X}_p\in \mathbb{R}^{L+T}$, where $T$ represents the forecasting horizon. Subsequently, $\mathbf{X}_p$ is  transformed into the frequency domain using the real-valued Fast Fourier Transform (rFFT), yielding $\mathbf{F}\in \mathbb{C}^{\lfloor\frac{L+T}{2}\rfloor+1}$. After being processed by Frequency Filters (F-Filter), we apply Frequency Reversible Instance Norm (F-RIN) to $\mathbf{F}$ and then feed it into the Complex-valued Spectrum Attention (CSA) block. Finally, we use the inverse real-valued Fast Fourier Transform (irFFT) to transform it back to the time domain and obtain the forecasting results.
In the subsequent part, we will provide a detailed description of the components of \model.

\subsection{Learnable Linear Padding}

\textbf{Frequency Resolution Alignment.} \model\ is designed to forecast the frequency spectrum of the complete series (both the input part and the forecasting part) based on the spectrum of the input series. However, the frequency resolution is determined by the length of the series. As a result, the resolutions of the two spectra are misaligned.
As shown in Figure \ref{fig:FreqMis}, when the resolutions of the spectra are misaligned, it becomes difficult to retrieve the corresponding information at specific frequencies of the complete-series spectrum based on the input spectrum.


The following is a detailed explanation of the Frequency Resolution Alignment problem.
According to the formula of DFT: $$\mathbf{F}[k]=\sum_{n=0}^{L-1}\mathbf{X}[n]e^{-2\pi i\frac{kn}{L}}, k=0,1,\dots,L-1,$$ each value in the DFT spectrum represents the inner product between the series ($\mathbf{X}[n]$) and a specific trigonometric basis ($e^{-2\pi i\frac{kn}{L}}$) at a corresponding frequency ($ \frac{2\pi k}{L}$).
For input series $\mathbf{X}_{inp}\in\mathbb{R}^{L}$ and complete series $\mathbf{X}_{comp}\in\mathbb{R}^{L+T}$, where $L$ is the look-back window and $T$ is the forecasting horizon, we denote the corresponding DFT spectrum as $\mathbf{F}_{inp}\in\mathbb{C}^{L}$, $\mathbf{F}_{comp}\in\mathbb{C}^{L+T}$.
In the case of $\mathbf{F}_{inp}$, the frequencies are denoted as $\mathit{f_{inp}}\in \{\frac{2\pi k}{L} | k=0,1,\dots,L-1\}$, and for $\mathbf{F}_{comp}$, the frequencies are $\mathit{f_{comp}}\in \{\frac{2\pi k}{L+T} | k=0,1,\dots,L+T-1\}$.

The misalignment of frequencies refers to a situation where the frequencies in $\mathit{f_{inp}}$ and $\mathit{f_{comp}}$ do not perfectly correspond. In other words, there may be important frequencies present in $\mathit{f_{comp}}$ that are not captured in $\mathit{f_{inp}}$.

To illustrate this, let's consider the example with an evident period of $24$, where $L=300$ and $T=60$, as depicted in Figure \ref{fig:FreqMis}. Our goal is to predict $\mathbf{F}_{comp}[k]$ using $\mathbf{F}_{inp}$, and we focus on the specific period $24$, which corresponds to the frequency index $15$ in the spectrum of complete series. The ground truth value for $\mathbf{F}_{comp}[15]$ is determined by summing the inner products between $\mathbf{X}_{comp}$ and the trigonometric basis at the corresponding frequency $\frac{\pi}{12}$ ($\frac{2\pi \times 15}{300 + 60}$).
However, when we examine the input spectrum $\mathbf{F}_{inp}$, we find the frequency $\frac{\pi}{12}$ is not present in $\mathit{f_{inp}}$. The two most adjacent frequencies in $\mathit{f_{inp}}$ are $\frac{2\pi}{25}$ and $\frac{13\pi}{150}$(corresponding to $k=12, 13$ respectively). This means the dominant period $24$ (corresponding to the frequency $\frac{\pi }{12}$) cannot be accurately detected in the spectrum of input series due to the misalignment of frequencies.

\textbf{Learnable Linear Padding.} To tackle the frequency-resolution misalignment problem, we propose the Learnable Linear Padding (LLP) method. LLP pads an input series of length $L$ to match the length of complete series $L+T$ with a learnable linear layer. This can be formulated as:
$$\mathbf{X}_p=\textbf{concat}(\mathbf{X}, \mathbf{W}^T\mathbf{X}+b),$$
where $\mathbf{X}\in \mathbb{R}^{L}$ represents the input series, $\mathbf{W}\in \mathbb{R}^{L\times T}$ and $b\in \mathbb{R}^{T}$  are the weight matrix and bias vector respectively, and $\mathbf{X}_p\in \mathbb{R}^{L+T}$ is the padded series.

By applying Learnable Linear Padding, we can obtain the spectrum of the input series with a frequency resolution that is aligned with that of the complete series. This effectively eliminates the frequency-resolution misalignment issue. As a result, it enables more effective extraction of characteristics from the frequency domain, ultimately contributing to enhanced forecasting performance.

\subsection{{Frequency Filter}}\label{sec:f-filter}

\begin{figure}[!htp]
\centering
\includegraphics[width=0.5\textwidth]{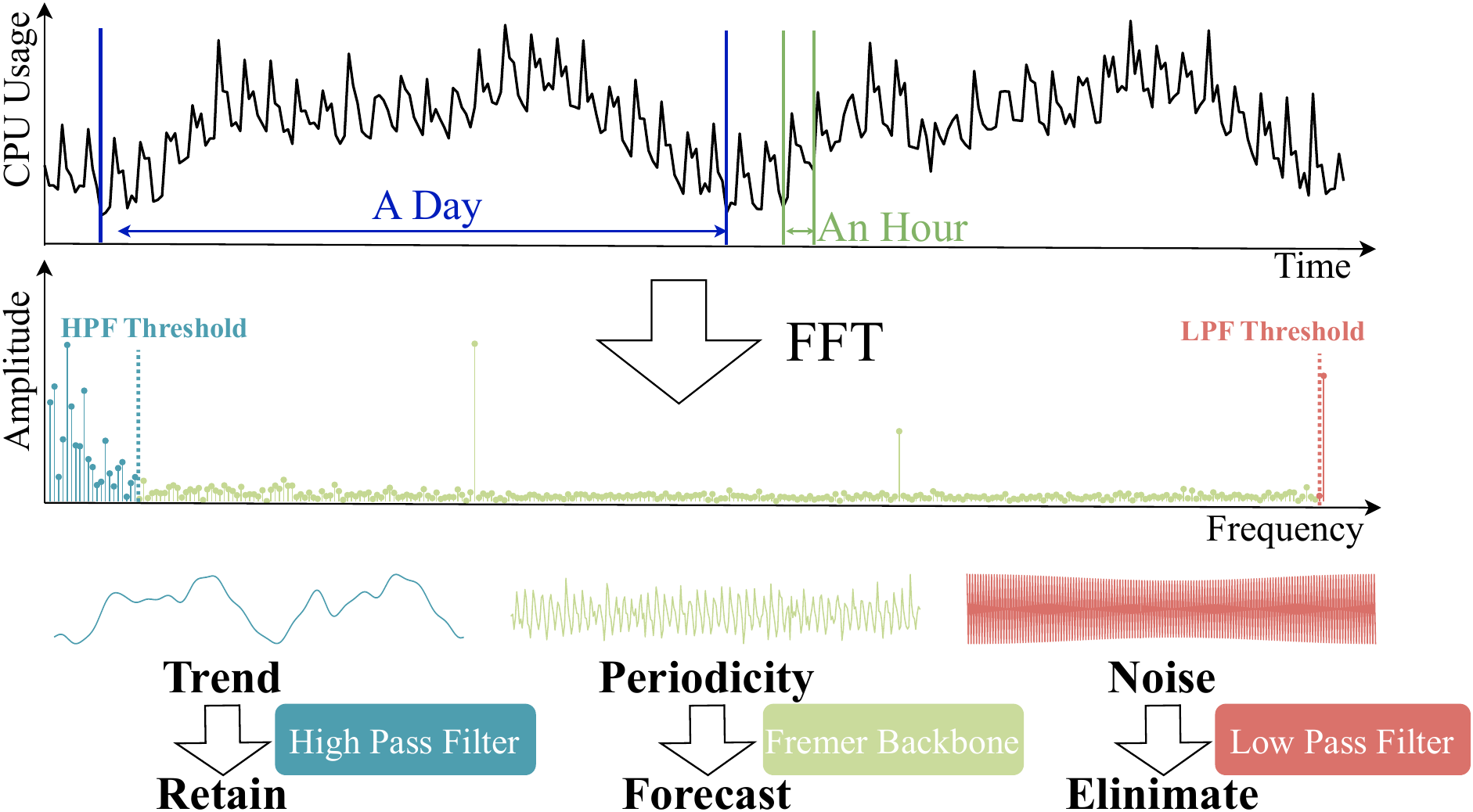}
\centering
\caption{{Frequency Filter is used to retain low-frequency trend information and eliminate high-frequency noise.}}
\label{fig:Filter}
\vspace{-6mm}
\end{figure}

{After obtaining the frequency domain representation with Frequency Resolution Alighment and Learnable Linear Padding, the model is tasked with learning effective global information. However, two primary technological challenges need to be addressed, as shown in Figure~\ref{fig:Filter}. First, the noise, especially high-frequency noise, inherent in the frequency domain introduces artifacts that degrade model performance~\citep{jiang2023fecam, yi2024filternet}. Second, deep learning approaches are inclined to prioritize low-frequency information during training due to its typically larger magnitude~\citep{piao2024fredformer}, which can result in the loss of critical periodic patterns. To address them, we design Low-Pass Filter (LPF) and High-Pass Filter (HPF) respectively.}

{LPF aims to eliminate noise by setting all frequency components above the LPF threshold (the highest 1\% of frequencies by default) to 0+0j. For the low-frequency components, two key observations are noted: First, the low-frequency portion typically contains trend information, such as the zero-frequency component, which captures the mean value of the sequence. Second, low-frequency components are a primary contributor to model overfitting, as the model prioritizes fitting these components first~\cite{piao2024fredformer}. We need both to retain the trend information in the low-frequency part and to prevent the model from overfitting to it. Consequently, we retain the low-frequency part (the lowest 3\% of frequencies by default) but exclude this information from \model backbone input to mitigate overfitting risks. After the \model backbone outputs predictions, we directly concatenate the low-frequency components with them to form the complete prediction results.}



\begin{figure*}[t]
\centering
\includegraphics[width=\textwidth]{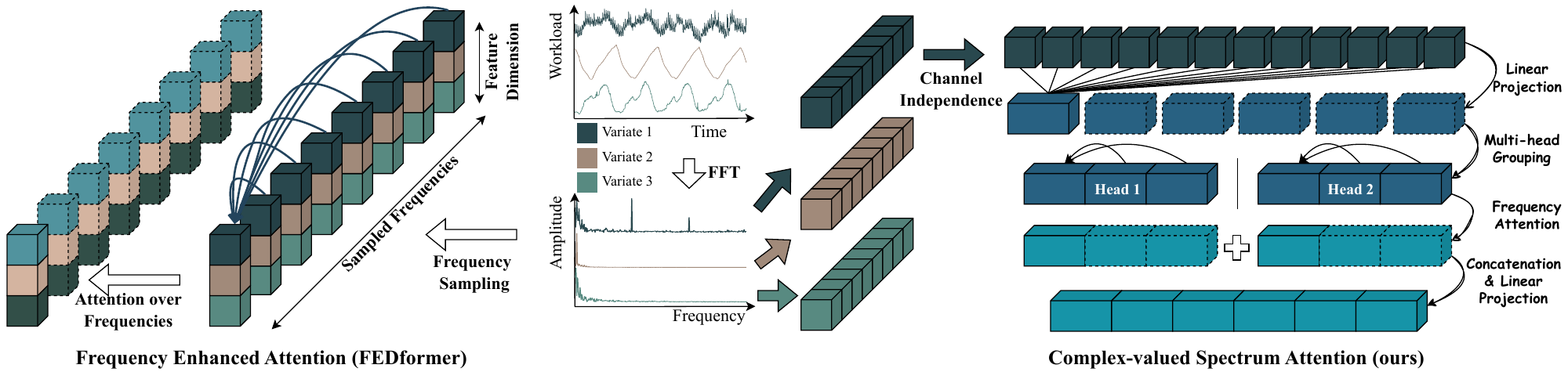}
\centering
\caption{ {Comparison between Frequency Enhanced Attention (FEA) proposed by FEDformer~\citep{zhou2022fedformer}) and our proposed CSA}.
}
\label{attention}
\vspace{0pt}
\end{figure*}
\subsection{Complex-valued Spectrum Attention}
\label{CSA}

After padded with LLP, transposing to frequency domain with rFFT and applying the frequency filters, we obtain the input spectrum $\mathbf{F}_f\in\mathbb{C}^{\lfloor\frac{L+T}{2}\rfloor+1-L_f}$, where  $L_f$ is the number of frequencies that are filtered out by the filters. 

Firstly, we utilize Frequency Reversible Instance Norm (F-RIN)  to process the input spectrum $\mathbf{F}_f$. While RevIN\citep{kim2021reversible} was originally developed to address distribution shift in the time domain, we have found it to be effective in handling frequency domain spectra as well. This approach enables transforming the spectra of series with distinct global features into a similar distribution. To accomplish this, $\mathbf{F}_f$ is normalized by subtracting its mean and dividing it by its standard deviation. The standard deviation is calculated using $|\mathbf{F}_f|\in\mathbb{R}^{\lfloor\frac{L+T}{2}\rfloor+1-L_f}$. Specifically, we obtain the normed spectrum $\mathbf{F}_r$ by the following process:
\begin{equation}
    \mathbf{F}_r = \frac{\mathbf{F}_f-\textbf{mean}(\mathbf{F}_f)}{\textbf{std}(|\mathbf{F}_f|)}.
\end{equation}

Afterwards, we add them back at the end of the structure. Next, a trainable linear projection $\mathbf{W}_{proj}\in \mathbb{C}^{L^\prime\times (\lfloor\frac{L+T}{2}\rfloor+1-L_f)}$ is employed to obtain $\mathbf{F}_c\in\mathbb{C}^{L^\prime}$, which represents the linear combinations of $\mathbf{F}_r$.

\textbf{Complex-valued Spectrum Attention}. 
As a single frequency point is scarcely capable of carrying semantic meaning, we developed Complex-valued Spectrum Attention (CSA). CSA is designed to capture attention across  frequency combinations, which enables it to more effectively capture the characteristics of the frequency domain.
Specifically, We apply a modified multi-head attention mechanism rather than the original one in \citep{vaswani2017attention}. For each head $h=1,2,\dots,H$, it projects the linear combined spectrum $\mathbf{F}_c\in\mathbb{C}^{L^\prime}$ to dimension $l$ across the spectrum dimension with trainable projections.
Specifically, $\mathbf{Q}_h=\mathbf{F}_c^T\mathbf{W}_h^Q$, $\mathbf{K}_h=\mathbf{F}_c^T\mathbf{W}_h^K$, $\mathbf{V}_h=\mathbf{F}_c^T\mathbf{W}_h^V$, where $\mathbf{W}_h^Q, \mathbf{W}_h^K, \mathbf{W}_h^V\in \mathbb{C}^{L^\prime\times l}$ are learnable parameters.
On each head we perform complex-valued Dot-Product Attention:
\begin{equation}
    \begin{aligned}
        \textbf{Attention}(\mathbf{Q}_h, \mathbf{K}_h, \mathbf{V}_h)&=\textbf{Softmax} (|\mathbf{Q}_h\mathbf{K}_h^T|)\mathbf{V}_h.\\
    \end{aligned}
\end{equation}

 Then the final output of Complex-valued Spectrum Attention is calculated as follows:
\begin{equation}
    \begin{aligned}
        \text{head}_h &= \textbf{Attention}(\mathbf{Q}_h, \mathbf{K}_h, \mathbf{V}_h),\\
        \text{CSA}(\mathbf{F}_c)&=\textbf{Concat}(\text{head}_1, \dots, \text{head}_H)^T\mathbf{W}_O,
    \end{aligned}
\end{equation}
where $\mathbf{W}_O\in \mathbb{C}^{hl\times L^\prime}$ are learnable parameters. The inverted multi-head projection not only enables capturing relationships among frequency combinations from multiple perspective, but also reduces the complexity to $\frac{1}{H}$ of the original approach.

Figure \ref{attention} presents a detailed visualization comparing the Complex-valued Spectrum Attention (CSA) with the Frequency Enhanced Attention (FEA) proposed in FEDformer \citep{zhou2022fedformer}, both of which employ attention in the frequency domain representation. There exist two key distinctions between them:
\begin{itemize}
    \item FEA employs a channel-mixing strategy, while CSA adopts a channel-independent approach. Since frequency domain representations primarily capture global patterns within a series—patterns unlikely to correlate across channels—we argue that channel independence is more effective for handling frequency domain data.
    \item FEA computes attention over sampled frequencies, while CSA computes attention over combinations of frequencies. Since a single frequency point is unlikely to carry semantic meaning, capturing attention across combinations of frequencies will yield greater benefits.
\end{itemize}


We follow the design of LayerNorm and FeedForward layers with residual connections in Transformer, with expanding to complex number field. 
Subsequently, following operations through a linear projection layer, a frequency-recovery process, and a Frequency Reversible Instance Norm (F-RIN) process, we obtain the output $\mathbf{\hat{F}}\in\mathbb{C}^{\lfloor\frac{L+T}{2}\rfloor+1}$.
We convert $\mathbf{\hat{F}}$ to time domain using inverse real-valued Fast Fourier Transform (irFFT), taking the last $T$ points (the forecasting part) as the final output of \model\ $\mathbf{\hat{X}}\in\mathbb{R}^T$.

\subsection{Complexity Analysis}\label{sec:Complexity}

{The primary computational bottleneck of the Transformer-based model lies in the Attention mechanism. The dot-product Attention mechanism has a complexity of $O(L²)$, where $L$ is the input sequence length. To reduce this complexity, \model uses a series of designs, ultimately lowering it to $O(\frac{L'^2}{H})$, where $L', H$ are the number of the frequency combinations and the Attention heads.}

{Firstly, according to the symmetry property of the Discrete Fourier Transform, the spectrum length in the frequency domain is $\frac{L}{2}$ for a sequence of length $L$ transformed into the frequency domain. Considering the use of LLP and Frequency Filter, the actual length is $\frac{L+T}{2}-L_f$. Furthermore, based on the observation that frequency-domain information is concentrated and exists in combinations, we map frequency information to combinations to further reduces computational complexity. The sequence length is further reduced to $L'$ (defaulting to $\frac{L}{5}$). The Attention formula is as follows:}
{
\begin{equation*}
    \textbf{Attention}(\mathbf{Q}, \mathbf{K}, \mathbf{V})=\textbf{Softmax} (|\mathbf{Q}\mathbf{K}^T|)\mathbf{V},
\end{equation*}
where $\mathbf{Q}=\mathbf{F}_c^T\mathbf{W}^Q$, $\mathbf{K}=\mathbf{F}_c^T\mathbf{W}^K$, $\mathbf{V}=\mathbf{F}_c^T\mathbf{W}^V$ and $\mathbf{Q}, \mathbf{K}, \mathbf{V}\in \mathbb{C}^{1\times L^\prime}$. Consequently, the computational complexity remains $O((L^{\prime})^2)$.
By applying the inverted multi-head projection, we decompose the original frequency combinations into $H$ independent projection heads: $\mathbf{Q}_h, \mathbf{K}_h, \mathbf{V}_h, \text{for } h=1\dots H$, and the attention computation for each head is then given by:
\begin{equation*}
    \textbf{Attention}(\mathbf{Q}_h, \mathbf{K}_h, \mathbf{V}_h)=\textbf{Softmax} (|\mathbf{Q}_h\mathbf{K}_h^T|)\mathbf{V}_h,
\end{equation*}
where $\mathbf{Q}_h=\mathbf{F}_c^T\mathbf{W}_h^Q$, $\mathbf{K}_h=\mathbf{F}_c^T\mathbf{W}_h^K$, $\mathbf{V}_h=\mathbf{F}_c^T\mathbf{W}_h^V$ and $\mathbf{Q}_h, \mathbf{K}_h, \mathbf{V}_h\in \mathbb{C}^{1\times \frac{L^\prime}{h}}$. The computational complexity will  be  $O(H\times(\frac{L^{\prime}}{H})^2)=O(\frac{(L^\prime)^2}{H})$. Compared to the dot-product Attention, the computational complexity of the Complex-valued Spectrum Attention used in \model is reduced to approximately $\frac{1}{200}$ of the original.
We further compare the attention complexity of \model\ with other Transformer-based forecasting models, as summarized in Table \ref{tab:complexity}.  
}

\begin{table}[htbp]
\centering
\caption{Attention Complexity of Transformer-based Forecasting Models. Let $L$ denote the input series length, $L^\prime$ the number of frequency combinations, $H$ the number of attention heads, $D$ the hidden dimension, $P$ the patching stride, and $L_f$ the number of sampled frequencies in FEDformer.}
\label{tab:complexity}
    \begin{tabular}{cc|cc}
    \toprule
    Model & Complexity & Model & Complexity \\
    \midrule
    Fremer & $O(\frac{{L^\prime}^2}{H})$ & {iTransformer} & {$O(D^2)$} \\
     Informer & $O(DL \log L)$ & {Fredformer} & {$O(\frac{D^2}{P}L)$} \\
    PatchTST & $O(\frac{D}{{P}^2}L^2)$ & FEDformer & $O(DL_f^2)$\\
    Transformer & $O(DL^2)$ &  Crossformer & $O(\frac{D}{{P}^2}L^2)$  \\
    \bottomrule
    \end{tabular}
\end{table}


\section{Experiments}\label{sec:exp}

{In this section, we conduct extensive experiments to address the following questions:}
\begin{itemize}
    \item[\textbf{RQ1}] {How does the performance of \model compare with SOTA time series forecasting models in workload forecasting?}
    \item[\textbf{RQ2}] {Can the key components within \model be identified as significant contributors to its performance?}
    \item[\textbf{RQ3}] {How well does \model deal with the historical data sparsity problem?}
    \item[\textbf{RQ4}] {Can \model well balance the efficiency and effectiveness compared to other models?}
    \item[\textbf{RQ5}] {Besides workload forecasting, can \model demonstrate competitive performance in other forecasting scenarios?}
    \item[\textbf{RQ6}] {In the cloud service scenarios, can \model help improve the system performance, such as achieving lower latency?}
\end{itemize}

\subsection{Experiment Settings}\label{sec:expsetting}
\textbf{Datasets.} We present four high-caliber workload datasets sourced from diverse cloud services within ByteDance, as detailed below:

\textbf{FaaS} (Function as a Service). ByteFaaS provides a highly scalable and efficient way to execute discrete functions. The dataset includes the QPS (Queries Per Second) data of function instances.

\textbf{IaaS} (Infrastructure as a Service). IaaS supplies users with foundational computing resources such as virtual machines, storage, etc.. The dataset records the CPU usage data of virtual machines.

\textbf{PaaS} (Platform as a Service). PaaS offers a comprehensive development and deployment environment. The dataset contains CPU usage information of individual services measured in milli-cores.

\textbf{RDS} (Relational Database Service). RDS is designed for efficient management and storage of structured data. The dataset includes QPS data from MySQL instances.

In addition, we also incorporate 3 public datasets from Materna-Workload-Traces\footnote{\href{https://www.kaggle.com/datasets/kpiyush04/maternaworkloadtraces}{https://www.kaggle.com/datasets/kpiyush04/maternaworkloadtraces}} \citep{urul2018energy}. These datasets are provided by GWA-T-13 MATERNA and originate from a German-based distributed cloud data center. They comprise three distinct traces, with each trace containing 520, 527, and 547 virtual machines (VMs), respectively. We choose the metric "CPU usage" as workload for forecasting and VMs exhibiting a missing rate (proportion of missing time points over total length) exceeding 2\% are filtered out.
We also summarize the information of datasets utilized in this study in Table \ref{dataset}.

\begin{table}[htbp]
  \caption{Statistics of workload datasets. }
  \label{dataset}
  \centering
  \begin{threeparttable}
  \renewcommand{\multirowsetup}{\centering}
  \setlength{\tabcolsep}{4.0pt}
  \begin{tabular}{c|c|c|c|c|c}
    \toprule
    {Dataset} & 
    {\rotatebox{0}{\scalebox{0.8}{\# of Instances}}} &
    {\rotatebox{0}{\scalebox{0.8}{{Lengths}}}} &
    {\rotatebox{0}{\scalebox{0.8}{Frequency}}} &
    {\rotatebox{0}{\scalebox{0.8}{Start Date}}} &
    {\rotatebox{0}{\scalebox{0.8}{End Date}}}  \\
    \midrule
{{\scalebox{0.95}{FaaS}}} 
&\scalebox{0.78}{226} &\scalebox{0.78}{2305} &\scalebox{0.78}{10-min} &\scalebox{0.78}{2022-04-02} &\scalebox{0.78}{2022-04-18}  \\
\midrule
{{\scalebox{0.95}{IaaS}}} 
&\scalebox{0.78}{93} &\scalebox{0.78}{3456} &\scalebox{0.78}{10-min} &\scalebox{0.78}{2023-06-30} &\scalebox{0.78}{2023-07-24} \\
\midrule
{{\scalebox{0.95}{PaaS}}} 
&\scalebox{0.78}{426} &\scalebox{0.78}{7776} &\scalebox{0.78}{10-min} &\scalebox{0.78}{2024-09-01} &\scalebox{0.78}{2024-10-24} \\
\midrule
{{\scalebox{0.95}{RDS}}} 
&\scalebox{0.78}{1113} &\scalebox{0.78}{6624} &\scalebox{0.78}{10-min} &\scalebox{0.78}{2024-08-16} &\scalebox{0.78}{2024-09-30} \\
\midrule
{{\scalebox{0.95}{MT-1}}} 
&\scalebox{0.78}{413} &\scalebox{0.78}{8352} &\scalebox{0.78}{5-min} &\scalebox{0.78}{2015-11-05} &\scalebox{0.78}{2015-12-03} \\
\midrule
{{\scalebox{0.95}{MT-2}}} 
&\scalebox{0.78}{402} &\scalebox{0.78}{8928} &\scalebox{0.78}{5-min} &\scalebox{0.78}{2015-12-04} &\scalebox{0.78}{2016-01-03} \\
\midrule
{{\scalebox{0.95}{MT-3}}} 
&\scalebox{0.78}{371} &\scalebox{0.78}{10368} &\scalebox{0.78}{5-min} &\scalebox{0.78}{2016-01-04} &\scalebox{0.78}{2016-02-08} \\
    
    \bottomrule
  \end{tabular}
  \end{threeparttable}
  \vspace{-1mm}
\end{table}


\begin{table*}[t]
  \caption{Results of Workload forecasting. The best results are bold and the second-best results are underlined. "Seasonality" and "Correlation" are statistical characteristics of the dataset calculated per TFB~\cite{qiu2024tfb}.}
  \label{tab:result_main}
  \centering
  \begin{threeparttable}
  \begin{small}
  \renewcommand{\multirowsetup}{\centering}
  \setlength{\tabcolsep}{1pt}

  \begin{tabular}{cccc|cccccccc|ccc|cc|c}
    \toprule
\multicolumn{4}{c|}{Architectures} &  \multicolumn{8}{c|}{Transformer} &  \multicolumn{3}{c|}{MLP} & \multicolumn{2}{c|}{CNN} & \multicolumn{1}{c}{Attention} \\
{\scalebox{0.8}{Datasets}} & 
{\scalebox{0.8}{{Seasonality}}} & 
{\scalebox{0.8}{Correlation}} & 
{\scalebox{0.8}{Metrics}} & 
{\rotatebox{0}{\scalebox{0.8}{{Fremer}}}} &
{\rotatebox{0}{\scalebox{0.8}{{Fredformer}}}} &
{\rotatebox{0}{\scalebox{0.8}{{PDF}}}} &
{\rotatebox{0}{\scalebox{0.8}{{iTransformer}}}} &
{\rotatebox{0}{\scalebox{0.8}{{PatchTST}}}} &
{\rotatebox{0}{\scalebox{0.8}{{Crossformer}}}} &
{\rotatebox{0}{\scalebox{0.8}{{FEDformer}}}} &
{\rotatebox{0}{\scalebox{0.8}{{Informer}}}} &
{\rotatebox{0}{\scalebox{0.8}{{FITS}}}} &
{\rotatebox{0}{\scalebox{0.8}{{DLinear}}}} &
{\rotatebox{0}{\scalebox{0.8}{{NLinear}}}} &
{\rotatebox{0}{\scalebox{0.8}{{MICN}}}} &
{\rotatebox{0}{\scalebox{0.8}{{TimesNet}}}} &
{\rotatebox{0}{\scalebox{0.8}{{FECAM}}}} 
\\
\midrule

\multirow{3}{*}{{\scalebox{0.95}{PaaS}}}  & \multirow{3}{*}{{\scalebox{0.95}{0.974}}} & \multirow{3}{*}{{\scalebox{0.95}{0.897}}}
&  \scalebox{1}{MSE} &\scalebox{1}{\textbf{0.062}}  & 0.072 & \underline{0.065} & 0.072 &\scalebox{1}{0.082} &\scalebox{1}{0.220} &\scalebox{1}{2.262} &\scalebox{1}{0.249} &\scalebox{1}{{0.066}} &\scalebox{1}{0.067} &\scalebox{1}{0.068} &\scalebox{1}{0.079} & 0.553 & 0.066 \\
&  &  & \scalebox{1}{MAE} &\scalebox{1}{\textbf{0.137}}  & 0.161 & 0.146 & 0.155 &\scalebox{1}{0.183} &\scalebox{1}{0.316} &\scalebox{1}{1.191} &\scalebox{1}{0.355} &\scalebox{1}{{0.145}} &\scalebox{1}{0.146} &\scalebox{1}{0.149} &\scalebox{1}{0.167}  & 0.563 & \underline{0.142} \\
&  &  & \scalebox{1}{SMAPE(\%)} &\scalebox{1}{\textbf{5.329}}  & 6.527 & 5.878 & 6.461 &\scalebox{1}{7.928} &\scalebox{1}{13.982} &\scalebox{1}{49.226} &\scalebox{1}{15.655} &\scalebox{1}{{5.664}} &\scalebox{1}{5.739} &\scalebox{1}{5.943} &\scalebox{1}{6.971} & 25.170 & \underline{5.656} \\
\midrule

\multirow{3}{*}{{\scalebox{0.95}{FaaS}}} &
\multirow{3}{*}{{\scalebox{0.95}{0.909}}} & \multirow{3}{*}{{\scalebox{0.95}{0.740}}}
&  \scalebox{1}{MSE} &\scalebox{1}{\textbf{0.289}} &  0.431 & \underline{0.319} & 0.327  &\scalebox{1}{{0.324}} &\scalebox{1}{1.123} &\scalebox{1}{1.499} &\scalebox{1}{0.900} &\scalebox{1}{0.430} &\scalebox{1}{0.365} &\scalebox{1}{0.342} &\scalebox{1}{0.383} & 0.631 &  0.397 \\
&  &  & \scalebox{1}{MAE} &\scalebox{1}{\textbf{0.314}}  & 0.402 & \underline{0.341} & 0.346  &\scalebox{1}{{0.350}} &\scalebox{1}{0.849} &\scalebox{1}{0.960} &\scalebox{1}{0.684} &\scalebox{1}{0.472} &\scalebox{1}{0.391} &\scalebox{1}{0.355} &\scalebox{1}{0.392} & 0.570 &  0.413 \\
&  &   & \scalebox{1}{SMAPE(\%)} &\scalebox{1}{\textbf{9.240}}  & 11.6431 & \underline{9.917} & 10.207 &\scalebox{1}{{10.626}} &\scalebox{1}{24.832} &\scalebox{1}{27.784} &\scalebox{1}{20.180} &\scalebox{1}{14.661} &\scalebox{1}{11.264} &\scalebox{1}{10.658} &\scalebox{1}{11.450} &
17.120 & 12.095\\
\midrule

\multirow{3}{*}{{\scalebox{0.95}{RDS}}} & \multirow{3}{*}{{\scalebox{0.95}{0.879}}} & \multirow{3}{*}{{\scalebox{0.95}{0.707}}}
&  \scalebox{1}{MSE} &\scalebox{1}{\textbf{1.292}}  & 1.520 & 1.489 & \underline{1.333} &\scalebox{1}{1.774} &\scalebox{1}{5.636} &\scalebox{1}{3.853} &\scalebox{1}{3.967} &\scalebox{1}{1.481} &\scalebox{1}{1.625} &\scalebox{1}{{1.445}} &\scalebox{1}{2.034} & 3.555 &  1.585 \\
& &  & \scalebox{1}{MAE} &\scalebox{1}{\textbf{0.364}}  & 0.407 & 0.396 & \underline{0.388} &\scalebox{1}{0.431} &\scalebox{1}{0.897} &\scalebox{1}{0.962} &\scalebox{1}{0.766} &\scalebox{1}{{0.405}} &\scalebox{1}{0.420} &\scalebox{1}{0.415} &\scalebox{1}{0.468} & 0.770 & 0.389 \\
& &  & \scalebox{1}{SMAPE(\%)} &\scalebox{1}{\textbf{8.663}}  & 9.861 & 9.702 & {9.145} &\scalebox{1}{10.714} &\scalebox{1}{25.207} &\scalebox{1}{27.254} &\scalebox{1}{20.973} &\scalebox{1}{9.828} &\scalebox{1}{{9.810}} &\scalebox{1}{10.147} &\scalebox{1}{10.867}   
& 21.4237 & \underline{8.760}
\\
\midrule

\multirow{3}{*}{{\scalebox{0.95}{IaaS}}} & \multirow{3}{*}{{\scalebox{0.95}{0.819}}} & \multirow{3}{*}{{\scalebox{0.95}{0.742}}}
&  \scalebox{1}{MSE} &\scalebox{1}{\textbf{0.708}}  & 0.744 & 0.755 & 0.770 &\scalebox{1}{\underline{0.746}} &\scalebox{1}{1.106} &\scalebox{1}{0.999} &\scalebox{1}{0.936} &\scalebox{1}{0.746} &\scalebox{1}{0.753} &\scalebox{1}{0.818} &\scalebox{1}{0.785}  & 1.099 & 0.762  \\
&  &  & \scalebox{1}{MAE} &\scalebox{1}{\textbf{0.556}}  & 0.580 & 0.593 & 0.610 &\scalebox{1}{\underline{0.582}} &\scalebox{1}{0.782} &\scalebox{1}{0.746} &\scalebox{1}{0.689} &\scalebox{1}{0.588} &\scalebox{1}{0.589} &\scalebox{1}{0.618} &\scalebox{1}{0.609} & 0.763 & 0.591\\
&  &  & \scalebox{1}{SMAPE(\%)} &\scalebox{1}{\textbf{16.249}} & 16.933 & 17.025 & 16.705 &\scalebox{1}{{17.083}} &\scalebox{1}{22.521} &\scalebox{1}{21.741} &\scalebox{1}{19.990} &\scalebox{1}{17.199} &\scalebox{1}{\underline{16.944}} &\scalebox{1}{17.870} &\scalebox{1}{17.806} & 21.872 & 17.132\\
\midrule

\multirow{3}{*}{{\scalebox{0.95}{MT1}}}  & \multirow{3}{*}{{\scalebox{0.95}{0.635}}} & \multirow{3}{*}{{\scalebox{0.95}{0.731}}} 
&  \scalebox{1}{MSE} &\scalebox{1}{{15.408}}  & 15.992 & \underline{15.250} & \textbf{15.205} &\scalebox{1}{{15.654}} &\scalebox{1}{15.654} &\scalebox{1}{16.092} &\scalebox{1}{16.988} &\scalebox{1}{15.707} &\scalebox{1}{15.766} &\scalebox{1}{16.076} &\scalebox{1}{18.226} & 16.173  & 15.368 \\
&  &  & \scalebox{1}{MAE} &\scalebox{1}{\underline{0.433}}  & 0.466 & 0.447 & {0.435} &\scalebox{1}{{0.474}} &\scalebox{1}{0.531} &\scalebox{1}{0.750} &\scalebox{1}{0.641} &\scalebox{1}{0.485} &\scalebox{1}{0.479} &\scalebox{1}{0.484} &\scalebox{1}{0.845}  & 0.636 & \textbf{0.415}\\
&  &  & \scalebox{1}{SMAPE(\%)} &\scalebox{1}{\textbf{18.828}}  & 20.116 & 19.127 & \underline{18.829} &\scalebox{1}{21.024} &\scalebox{1}{24.734} &\scalebox{1}{38.943} &\scalebox{1}{26.474} &\scalebox{1}{{20.269}} &\scalebox{1}{21.636} &\scalebox{1}{20.696} &\scalebox{1}{33.932} & 27.567 & 19.152 \\
\midrule

\multirow{3}{*}{{\scalebox{0.95}{MT3}}}  & \multirow{3}{*}{{\scalebox{0.95}{0.628}}} & \multirow{3}{*}{{\scalebox{0.95}{0.723}}}
& \scalebox{1}{MSE} &\scalebox{1}{\textbf{28.062}}  & \scalebox{1}{32.778} & \underline{28.224} & 28.937 &\scalebox{1}{28.476} &\scalebox{1}{72.847} &\scalebox{1}{35.837} &\scalebox{1}{36.365} &\scalebox{1}{{28.391}} &\scalebox{1}{33.205} &\scalebox{1}{29.090} &\scalebox{1}{47.147} & 35.895 & 50.023\\
&  &  & \scalebox{1}{MAE} &\scalebox{1}{\underline{0.699}}  & 0.777  & \textbf{0.687 }& 0.713 &\scalebox{1}{\underline{0.718}} &\scalebox{1}{1.120} &\scalebox{1}{0.965} &\scalebox{1}{1.074} &\scalebox{1}{0.761} &\scalebox{1}{0.856} &\scalebox{1}{0.851} &\scalebox{1}{1.079} & 0.946 & 0.852 \\
&  &  &\scalebox{1}{SMAPE(\%)} &\scalebox{1}{\textbf{21.411}}  & 23.620 & 22.352 & 21.886 &\scalebox{1}{\underline{22.645}} &\scalebox{1}{25.859} &\scalebox{1}{34.105} &\scalebox{1}{33.722} &\scalebox{1}{22.911} &\scalebox{1}{24.333} &\scalebox{1}{29.269} &\scalebox{1}{29.703}  & 28.225 & 22.220 \\
\midrule

\multirow{3}{*}{{\scalebox{0.95}{MT2}}}  & \multirow{3}{*}{{\scalebox{0.95}{0.618}}}  & \multirow{3}{*}{{\scalebox{0.95}{0.714}}}
&  \scalebox{1}{MSE} &\scalebox{1}{{{6.543}}} & 6.618 & \underline{6.481} &  6.559 &\scalebox{1}{{\textbf{6.467}}} &\scalebox{1}{6.754} &\scalebox{1}{6.907} &\scalebox{1}{6.969} &\scalebox{1}{6.720} &\scalebox{1}{6.678} &\scalebox{1}{6.678} &\scalebox{1}{6.758} & 6.848 & 6.554 \\
&  &  & \scalebox{1}{MAE} &\scalebox{1}{\textbf{0.287}}  & 0.309 & 0.296 & \underline{0.290} &\scalebox{1}{{0.302}} &\scalebox{1}{0.356} &\scalebox{1}{0.532} &\scalebox{1}{0.447} &\scalebox{1}{0.352} &\scalebox{1}{0.332} &\scalebox{1}{0.305} &\scalebox{1}{0.401} & 0.433  & 0.300\\
& &   & \scalebox{1}{SMAPE(\%)} &\scalebox{1}{\textbf{14.360}}  & 16.038 & 14.629 & 14.927 &\scalebox{1}{15.470} &\scalebox{1}{19.329} &\scalebox{1}{31.562} &\scalebox{1}{20.659} &\scalebox{1}{16.064} &\scalebox{1}{17.628} &\scalebox{1}{\underline{14.554}} &\scalebox{1}{21.145} & 20.676  & 16.614\\

    \bottomrule
  \end{tabular}
    \end{small}
  \end{threeparttable}
\end{table*}


\textbf{Baselines.} {We select representative models as baselines, including Fredformer~\citep{piao2024fredformer}, PDF~\citep{dai2024periodicity}, iTransformer~\cite{liu2023itransformer}, FEDformer~\citep{zhou2022fedformer}, Crossformer~\citep{zhang2022crossformer}, Informer~\citep{zhou2021informer}, PatchTST~\citep{nie2022time},  DLinear~\citep{zeng2023transformers}, NLinear~\citep{zeng2023transformers}, FITS~\citep{xu2024fits}, MICN~\citep{wang2023micn}, TimesNet~\citep{wu2022timesnet}, and FECAM~\citep{jiang2023fecam}. }

\textbf{Evaluation Platform.} To ensure a fair and unbiased comparison, we adopt the Time Series Forecasting Benchmark (TFB)\citep{qiu2024tfb} as our evaluation platform. TFB includes implementations of all baseline models, and we retain the default settings for each model as provided by the platform. The datasets are systematically organized in the TFB format and divided into training, validation, and test sets using a 7:1:2 ratio based on the time span.

\textbf{Evaluation Metrics.} 
{Due to the magnitude differences across datasets (e.g., CPU utilization ranges from $[0,100]$, while QPS can reach up to $10^6\sim10^7$), computing metrics like mean squared error (MSE) and mean absolute error (MAE) on raw data is dataset-dependent and lacks comparability across different datasets. Therefore, we first normalize the data to calculate MSE$_{norm}$ and MAE$_{norm}$. Also, since symmetric mean absolute percentage error (SMAPE), which measures the relative magnitude of prediction errors, is not affected by the magnitude of the data, we compute it on raw data to comprehensively evaluate prediction errors. The specific calculating process of the test phase are as follows:
\begin{equation*}
\begin{aligned}
(\mathbf{X}_{\text{norm}}^{(n)},\mathbf{Y}_{\text{norm}}^{(n)}) &= \text{Normalization} ((\mathbf{X}^{(n)},\mathbf{Y}^{(n)})), \\
\mathbf{\hat{Y}}_{\text{norm}}^{(n)} &= \text{Fremer}(\mathbf{X}_{\text{norm}}^{(n)}), \\
\text{MSE}_{\text{norm}} &= \frac{1}{NH}\sum_{n=0}^{N-1}||\mathbf{\hat{Y}}_{\text{norm}}^{(n)}-\mathbf{Y}_{\text{norm}}^{(n)}||_2, \\
\text{MAE}_{\text{norm}} &= \frac{1}{NH}\sum_{n=0}^{N-1}||\mathbf{\hat{Y}}_{\text{norm}}^{(n)}-\mathbf{Y}_{\text{norm}}^{(n)}||_1, \\
\mathbf{\hat{Y}}^{(n)} &= \text{Denormalization}(\mathbf{\hat{Y}}_{\text{norm}}^{(n)}), \\
\text{SMAPE} &= \frac{100\%}{NH}\sum_{n=1}^{N}\sum_{i=1}^{H}\frac{2|\mathbf{\hat{Y}}_i^{(n)}-\mathbf{Y}_i^{(n)}|}{|\mathbf{\hat{Y}}_i^{(n)}|+| \mathbf{Y}_i^{(n)}|}, \\
\end{aligned}
\end{equation*}
where $N$ denotes the size of the test set, $H$ denotes the forecasting horizon, $(\mathbf{X}^{(n)}=[\mathbf{X}^{(n)}_1,\dots,\mathbf{X}^{(n)}_L],\mathbf{Y}^{(n)}=[\mathbf{Y}^{(n)}_1,\dots,\mathbf{Y}^{(n)}_H])$ denotes the $n$-th test sample, and $\mathbf{\hat{Y}}_{\text{norm}}^{(n)},\mathbf{\hat{Y}}^{(n)}\in\mathbb{R}^N$ denote the normalized and denormalized forecasts, respectively.
It should be emphasized that we use the sequence mean and standard deviation of the training set for data normalization and denormalization, thus avoiding test data leakage.}

\textbf{Implementation Details.} We implement \model\ and all baseline models utilizing the PyTorch framework \citep{paszke2019pytorch}, and conduct the training process on NVIDIA A100-SXM 80GB GPUs. The ADAM optimizer \citep{kingma2014adam} is selected, with an initial learning rate of $1e^{-3}$, and the  MSE-Loss is designated as the optimization objective. Throughout all experiments, the batch size is consistently set to 32, and the number of training epochs is fixed at 20.
Regarding the model hyper-parameters, such as the model dimension and the number of encoder layers, for the baseline models, we adopt the default settings provided within TFB. For \model, a limited hyper-parameter search is performed, with the primary focus on the frequency filter threshold. In the case of training hyper-parameters, including batch size and learning rate, a uniform configuration is applied across all models. This standardized implementation approach ensures the reproducibility and comparability of the experimental results, facilitating a more accurate assessment of the performance of \model\ and the baseline models within the research context.

\subsection{RQ1: Workload Forecasting Result}
\label{exp:forecasting}
In this section, an extensive assessment of the forecasting performance of \model\ in conjunction with other baseline models is conducted across the seven aforementioned workload datasets. The input window is configured as 5 days, corresponding to 1440 data points for datasets with a 5-minute granularity and 720 points for those with a 10-minute granularity. The forecasting horizon is set to 1 day, equivalent to 288 points for 5-minute granularity datasets and 144 points for 10-minute granularity datasets.

\begin{figure*}[!htbp]
    \centering     

        \includegraphics[width=\textwidth]{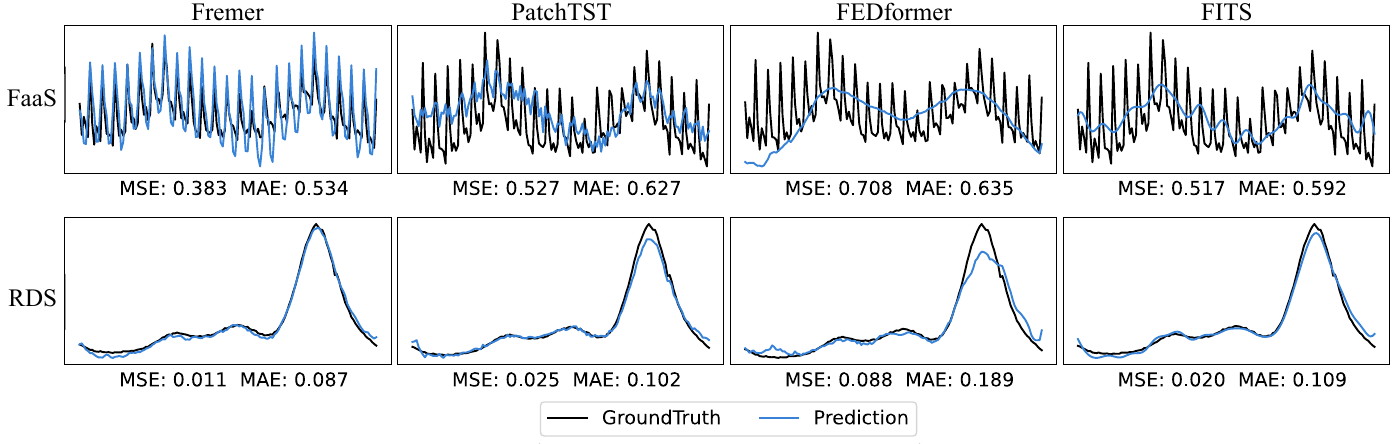}
    \vspace{-8mm}
    \caption{Visualization of Workload Forecasting Results on Distinct Datasets.}  
    \label{fig:vis}
\end{figure*}

\begin{figure*}[!htbp]
    \centering
        \includegraphics[width=\textwidth]{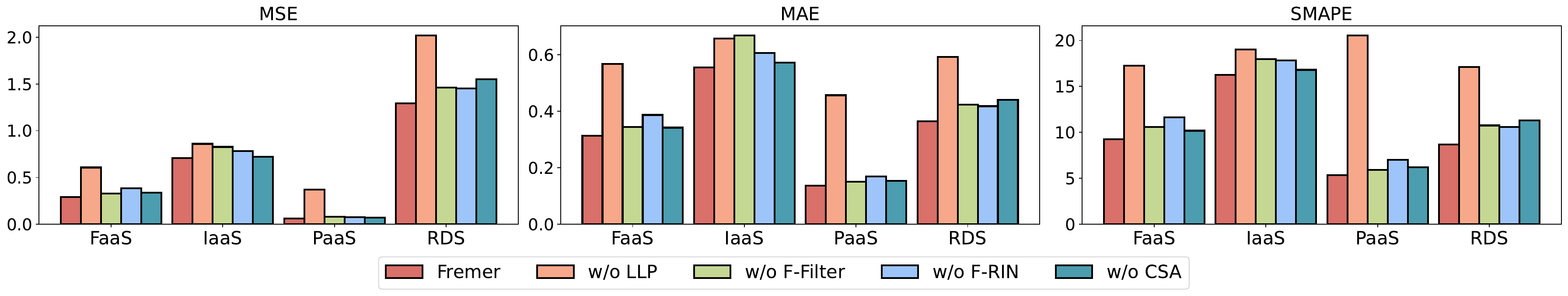}
    \vspace{-8mm}
    \caption{{Ablation Study Results. We highlight the contributions of four key modules of \model: Learnable Linear Padding (LLP), Frequency Filter (F-Filter), Frequency Instance Normalization (F-RIN), and Complex-valued Spectrum Attention (CSA).}}
    \label{fig:ablation_results}
\end{figure*}

As the results shown in Table \ref{tab:result_main}, \model\ achieve the best performance on most datasets. Specifically, in comparison to the best performing baseline models within each dataset, \model\ attains an average enhancement in forecasting accuracy of 2.9\% in terms of MSE, 2.5\% in MAE, and 3.5\% in SMAPE. These outcomes affirm the superiority of \model\ for workload forecasting. 
Additionally, it can be observed that for models that explicitly model the correlative relationships among workload series of distinct computing instances, their forecasting performance is relatively inferior to those models with channel independence. This phenomenon can be attributed to the relatively autonomous nature of workloads across distinct instances within these datasets, rendering the modeling of such correlative relationships ineffective. 


We present visualizations of the forecasting results of \model\ in comparison with some high-performing baseline models, as depicted in Figure \ref{fig:vis}. From these visualizations, it can be clearly observed that \model\ produces more accurate forecasts compared to other models. 
FaaS data shows complicated periodic pattern and \model\ captures the pattern accurately. RDS data shows local period shift and \model produce accurate predictions.

\subsection{RQ2: Components Analysis}

\subsubsection{Ablation Study.}
{We conduct the ablation study on the \model's key components, the Learnable Linear Padding (LLP), the Complex-valued Spectrum Attention (CSA), the Frequency Filter (F-Filter) and the Frequency-Reversible Instance Normalization (F-RIN) to analyze their contributions, as shown in Figure~\ref{fig:ablation_results}.
Specifically, the notations "w/o LLP", "w/o F-Filter", and "w/o F-RIN" designate \model\ without the corresponding modules, while "w/o CSA"  represents \model\ implemented with a frequency attention mechanism computed across individual frequency components.
We observe that each component plays a crucial role in enhancing the forecasting performance of \model. Specifically, removing the LLP module leads to a significant drop in performance, highlighting the critical importance of frequency alignment. By effectively using the LLP to align frequency resolutions, \model achieves higher forecasting accuracy. Besides, when the dataset, such as IaaS, has strong noise, removing F-Filter will incur the remarkable performance degradation.
Similarly, the absence of F-RIN and CSA will also cause the performance degradation, which underscores the effectiveness of applying RevIN to frequency domain representations and performing Attention on frequency combinations.}

\subsubsection{Frequency Filters}
\begin{figure}[!htbp]
    \centering     

        \includegraphics[width=0.5\textwidth]{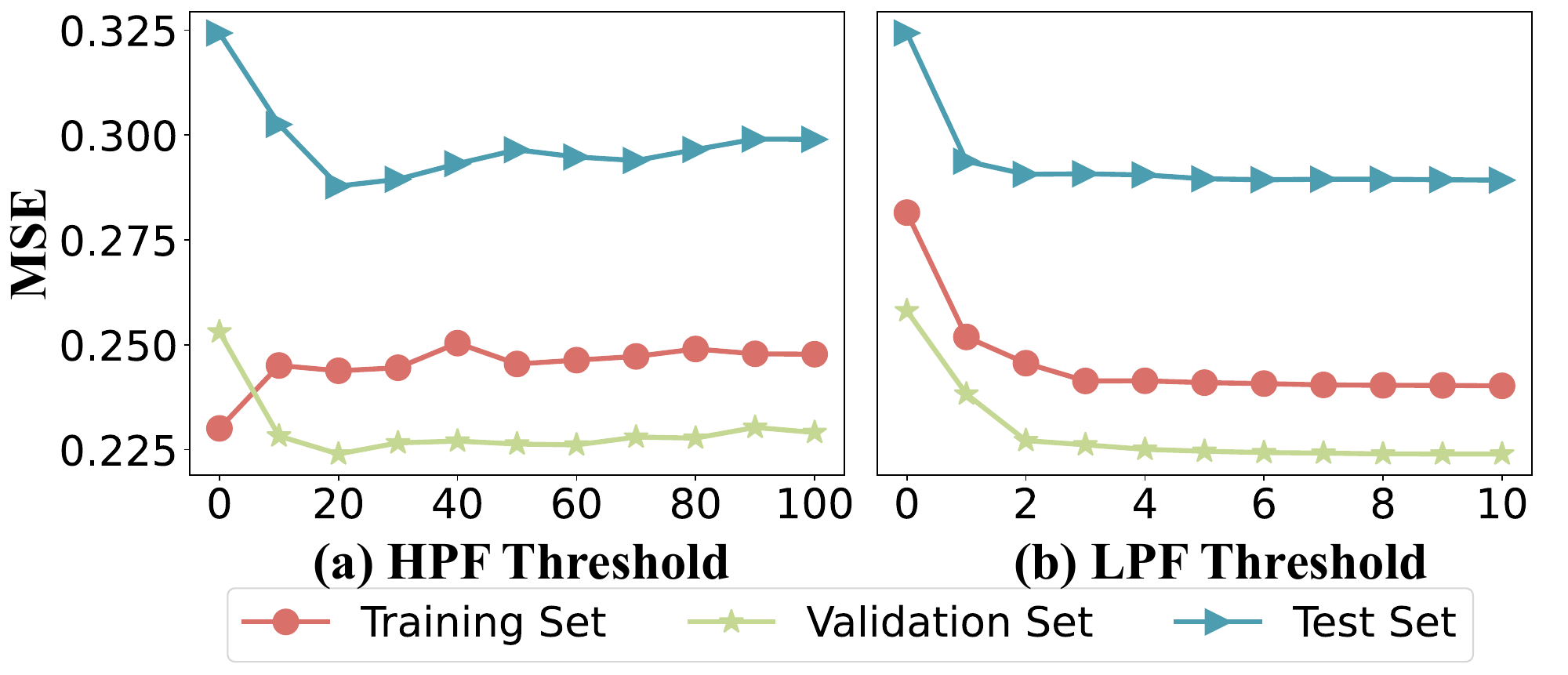}
    \caption{{Frequency Filter Analysis Results. Figure (a) and (b) respectively show the impact of different HPF and LPF thresholds on training/validation/testing losses.} }  
    \label{fig:filter1}
    \vspace{-4mm}
\end{figure}

{We also investigate the impact of frequency filter thresholds on \model. The results are presented in Figure~\ref{fig:filter1}. From Figure~\ref{fig:filter1}(a), it can be observed that when applying the High-Pass Filter (HPF), even with a small threshold, the training loss increases while the validation and test sets decrease. This indicates that without the HPF, \model\ has a tendency to overfit on the training set. Since the low-frequency part has a relatively larger influence on the forecasting results, simply retaining the low-frequency part can effectively address the overfitting problem. This provides a convenient way to control the generalizability of \model for different workloads. From Figure~\ref{fig:filter1}(b), it can be seen that removing the highest frequency noise reduces the training, validation, and testing losses. Only the highest-frequency noise significantly affects model performance, and carefully removing it can help the model learn better while retaining the useful information.}

\subsubsection{{Parameter Sensitivity}}\label{sec:sensitivily}

\begin{figure}[!htp]
    \centering     

        \includegraphics[width=0.48\textwidth]{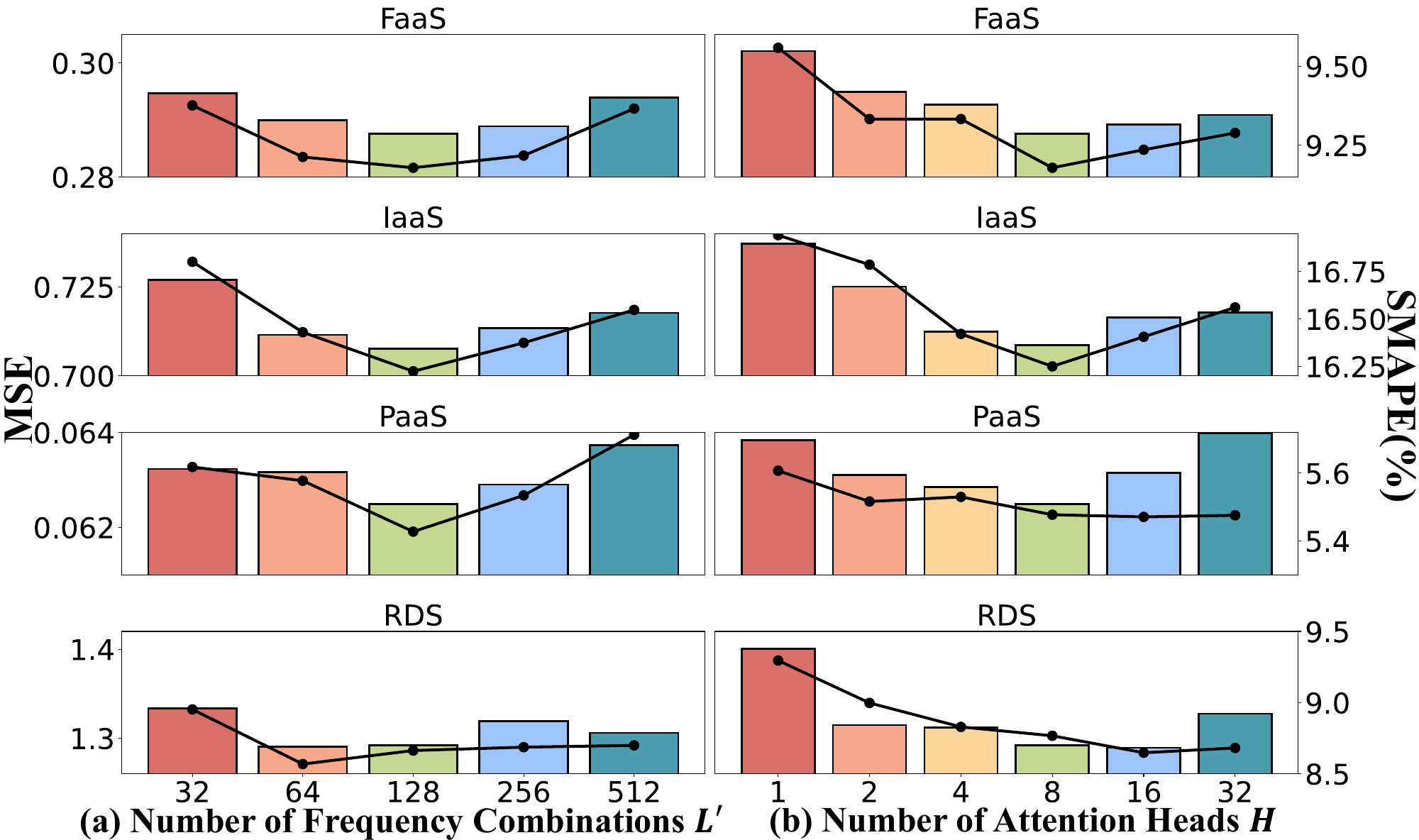}
    \caption{{The Parameter Analysis. The bar chart indicates MSE, and the line chart indicates SMAPE. Figure (a) shows the impact of the frequency combination number $L^\prime$, and (b) shows the impact of the attention head number $H$.}}  
    \label{fig:sensitivity}
\end{figure}

{We analyze the impact of key parameters on model performance, as shown in Figure~\ref{fig:sensitivity}. Figure~\ref{fig:sensitivity}(a) shows the impact of different frequency combination numbers $L'$. As can be seen, a larger $L'$ does not always lead to better results. When $L^\prime$ is approximately $\frac{L}{5}$ (for an input length of $L=720$, $L'=144$), \model achieves optimal performance. This aligns with our discussion in Section 4.4 and explains the efficiency of Fremer, as it can capture long sequence periodic features with fewer frequency combinations. Figure~\ref{fig:sensitivity}(b) shows the impact of different attention head numbers $h$. Overall, as $h$ increases, the performance of \model improves, but with diminishing returns. Excessively large $h$ can even increase error. Considering the balance between efficiency and effectiveness, $h=8$ is a reasonable choice.}

\subsubsection{{Input Length $L$ and Forecast Horizon $T$}}\label{sec:sensitivity_LH}
\begin{figure}[!htp]
    \centering     

        \includegraphics[width=0.48\textwidth]{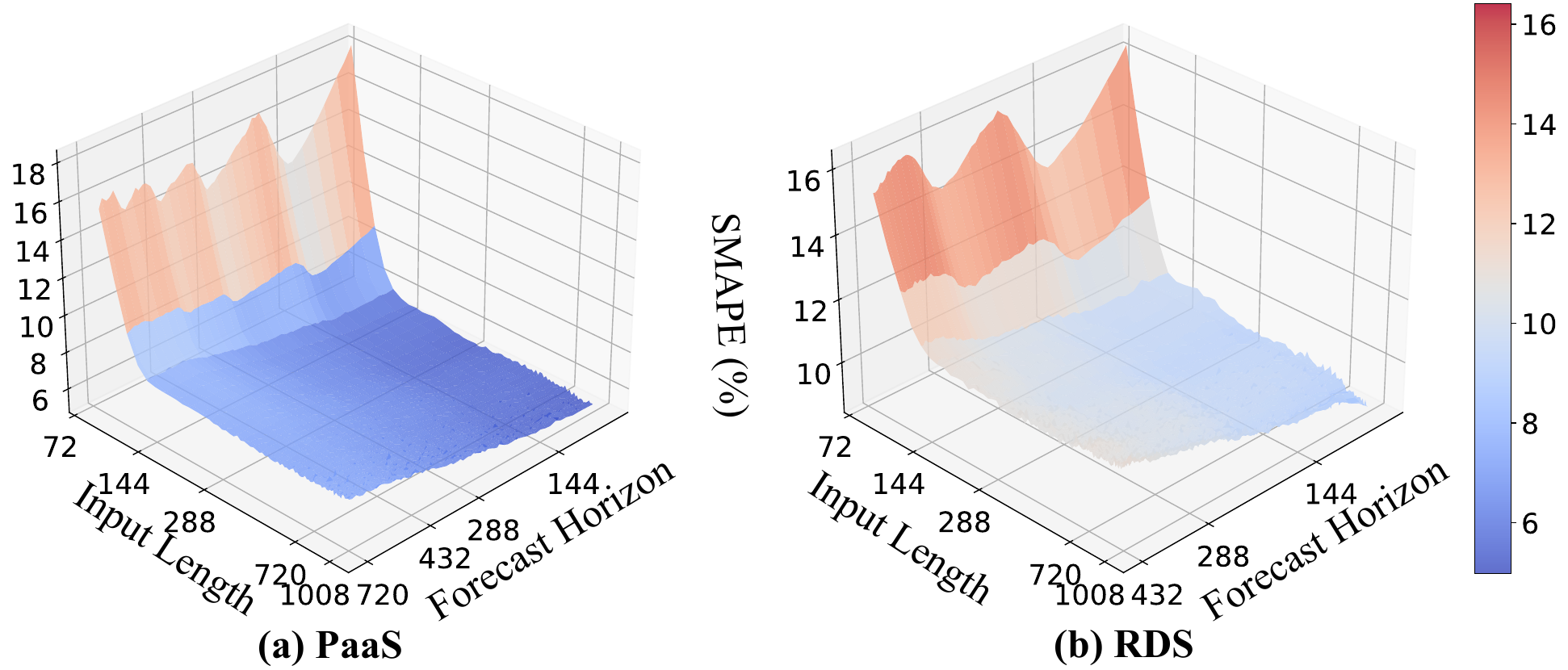}
    \caption{{The sensitivity to $L$ and $T$ of \model. The three axes respectively represent the input length $L$, the forecast horizon $T$, and the SMAPE.}}  
    \label{fig:sensitivity_LH}
\end{figure}

{We evaluate \model on the PaaS and RDS datasets across varying input lengths ($L \in [36,1008]$) and forecasting horizons ($H \in [18, 720]$, spanning hours to days), with results in Figure~\ref{fig:sensitivity_LH}. They show that once $L$ exceeds $144$ (i.e., the input includes a day’s data), \model’s prediction error drops sharply, highlighting its ability to capture data periodicity. The impact of output length on the model's effectiveness is influenced by the input length. When the input length is short ($L < 100$), the model's effectiveness is better when the output length is a multiple of the period. 
After the input length increases, the output results become more stable. When the input length is $720$, the SMPAE for an output length of $720$ increases by only 40\% compared to that for an output length of $72$.}

\subsubsection{{Multi-Head Attention Mechanism}}\label{sec:attention_visual}
\begin{figure}[!htp]
    \centering     
        \includegraphics[width=0.5\textwidth]{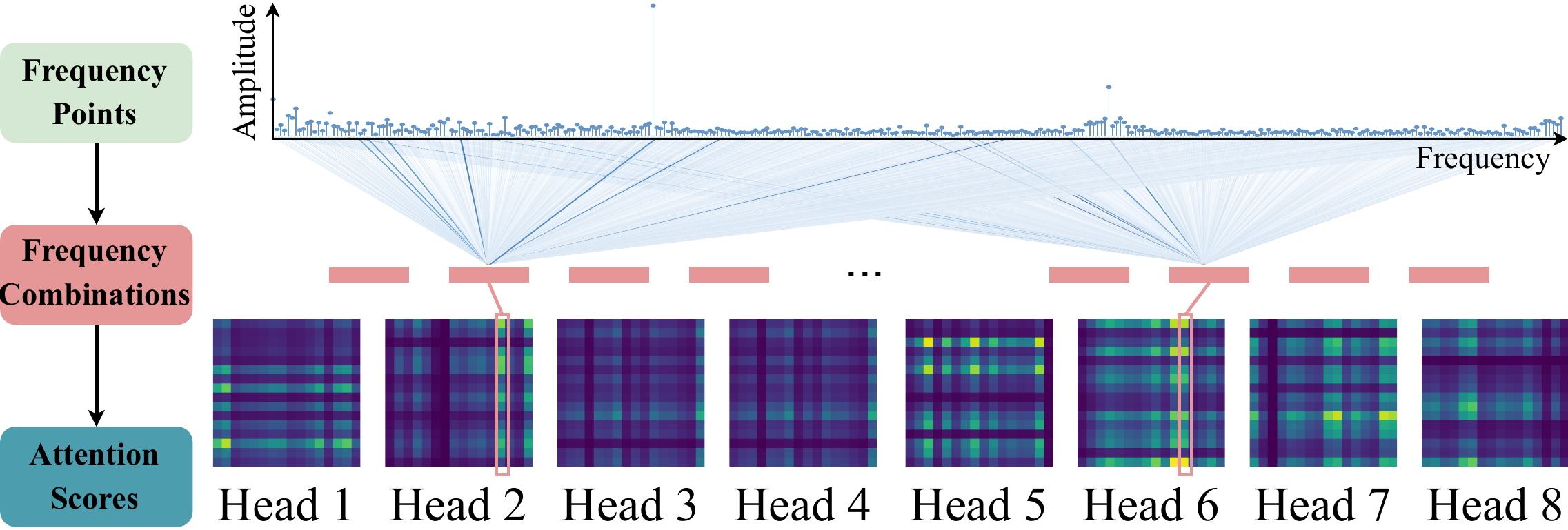}
    \caption{{Multi-head CSA. The figure shows the mapping from frequency points (shown in the top spectrum) to frequency combinations (indicated by middle red rectangles). The thicker the blue lines between them, the higher the weights. The bottom part is a heatmap of attention scores, where brightness indicates the score's magnitude. }}  
    \label{fig:multihead}
\end{figure}

{We visualize and analyze the multi-head CSA when forecasting FaaS workload, as shown in the figure~\ref{fig:multihead}. By embedding frequency points to combinations, CSA represents periodic information with lower complexity. The Multi-head mechanism further enhances the capture of different periodic elements. In the attention score heatmap, row $n$ shows the attention scores of frequency combination $m$ to others, and column $n$ shows the attention scores received by combination $n$ from others. In Head $2$ and Head $6$, the two most attended combinations are highlighted, showing their focus on different frequency information: the left on the hourly period, the right on the sub-hourly period.}

\begin{table}[t]
  \caption{Results of Intra-Dataset Transfer Forecasting. }
  \label{tab:in_transfer}
  \centering
  \begin{threeparttable}
  \begin{small}
  \renewcommand{\multirowsetup}{\centering}
  \setlength{\tabcolsep}{1pt}
  
  \begin{tabular}{c|c|ccccccc}
    \toprule
\multicolumn{2}{c}{} &  
{\rotatebox{0}{\scalebox{1.0}{{Fremer}}}} &
{\rotatebox{0}{\scalebox{1.0}{{PDF}}}} &
{\rotatebox{0}{\scalebox{1.0}{{iTrans.}}}} &
{\rotatebox{0}{\scalebox{1.0}{{PatchTST}}}} &
{\rotatebox{0}{\scalebox{1.0}{{FITS}}}} &
{\rotatebox{0}{\scalebox{1.0}{{DLinear}}}} &
{\rotatebox{0}{\scalebox{1.0}{{NLinear}}}}  \\
\midrule

\multirow{2}{*}{{\scalebox{0.95}{FaaS}}} 
&  \scalebox{0.98}{MSE} &\scalebox{0.98}{\textbf{0.325}}  & \scalebox{0.98}{0.335}  &   \scalebox{0.98}{0.355}      &\scalebox{0.98}{0.337} &\scalebox{0.98}{0.349} &\scalebox{0.98}{0.320} &\scalebox{0.98}{0.348}   \\
&  \scalebox{0.98}{SMAPE(\%)} &\scalebox{0.98}{\textbf{9.645}}  &  \scalebox{0.98}{9.866}   & \scalebox{0.98}{10.331 }         &\scalebox{0.98}{9.834} &\scalebox{0.98}{10.475} &\scalebox{0.98}{9.782} &\scalebox{0.98}{10.161}   \\
\midrule
\multirow{2}{*}{{\scalebox{0.95}{IaaS}}} 
&  \scalebox{0.98}{MSE} &\scalebox{0.98}{\textbf{0.707}}   & \scalebox{0.98}{0.710}   &  \scalebox{0.98}{0.714}         &\scalebox{0.98}{0.722} &\scalebox{0.98}{0.708} &\scalebox{0.98}{0.711} &\scalebox{0.98}{0.720}   \\
&  \scalebox{0.98}{SMAPE(\%)} &\scalebox{0.98}{{15.988}}  &  \scalebox{0.98}{16.023}   &  \scalebox{0.98}{16.270}         &\scalebox{0.98}{16.003} &\scalebox{0.98}{16.088} &\scalebox{0.98}{\textbf{15.979}} &\scalebox{0.98}{16.043}   \\
\midrule
\multirow{2}{*}{{\scalebox{0.95}{RDS}}} 
&  \scalebox{0.98}{MSE} &\scalebox{0.98}{\textbf{0.351}} &  \scalebox{0.98}{\textbf{0.351}}   &   \scalebox{0.98}{0.360}      &\scalebox{0.98}{0.388} &\scalebox{0.98}{0.371} &\scalebox{0.98}{0.361} &\scalebox{0.98}{0.368}   \\
&  \scalebox{0.98}{SMAPE(\%)} &\scalebox{0.98}{\textbf{9.688}}&  \scalebox{0.98}{9.695}   &   \scalebox{0.98}{10.232}        &\scalebox{0.98}{10.699} &\scalebox{0.98}{10.652} &\scalebox{0.98}{10.384} &\scalebox{0.98}{10.573}   \\
\midrule
\multirow{2}{*}{{\scalebox{0.95}{PaaS}}} 
&  \scalebox{0.98}{MSE} &\scalebox{0.98}{\textbf{0.045}}  & \scalebox{0.98}{ 0.048}  &   \scalebox{0.98}{0.051}         &\scalebox{0.98}{0.051} &\scalebox{0.98}{0.052} &\scalebox{0.98}{0.050} &\scalebox{0.98}{0.052}   \\
&  \scalebox{0.98}{SMAPE(\%)} &\scalebox{0.98}{\textbf{6.503}}   &  \scalebox{0.98}{6.619}   &  \scalebox{0.98}{6.582}         &\scalebox{0.98}{6.659} &\scalebox{0.98}{8.723} &\scalebox{0.98}{8.114} &\scalebox{0.98}{7.687}   \\
\midrule
\multirow{2}{*}{{\scalebox{0.95}{MT1}}} 
&  \scalebox{0.98}{MSE} &\scalebox{0.98}{{6.856}}  &  \scalebox{0.98}{7.080}   &  \scalebox{0.98}{7.033}         &\scalebox{0.98}{6.869} &\scalebox{0.98}{\textbf{6.798}} &\scalebox{0.98}{6.874} &\scalebox{0.98}{7.448}   \\
&  \scalebox{0.98}{SMAPE(\%)} &\scalebox{0.98}{\textbf{23.244}} & \scalebox{0.98}{23.595}   &   \scalebox{0.98}{24.207}         &\scalebox{0.98}{25.851} &\scalebox{0.98}{25.419} &\scalebox{0.98}{26.590} &\scalebox{0.98}{37.013}   \\
\midrule
\multirow{2}{*}{{\scalebox{0.95}{MT2}}} 
&  \scalebox{0.98}{MSE} &\scalebox{0.98}{\textbf{6.782}}  &  \scalebox{0.98}{6.982}   &   \scalebox{0.98}{6.835}        &\scalebox{0.98}{7.074} &\scalebox{0.98}{7.093} &\scalebox{0.98}{7.017} &\scalebox{0.98}{7.177}   \\
&  \scalebox{0.98}{SMAPE(\%)} &\scalebox{0.98}{\textbf{17.320}} & \scalebox{0.98}{17.500}   &  \scalebox{0.98}{18.319 }         &\scalebox{0.98}{20.602} &\scalebox{0.98}{18.810} &\scalebox{0.98}{18.151} &\scalebox{0.98}{17.534}   \\
\midrule
\multirow{2}{*}{{\scalebox{0.95}{MT3}}} 
&  \scalebox{0.98}{MSE} &\scalebox{0.98}{{15.378}}&   \scalebox{0.98}{15.221}   &   \scalebox{0.98}{ 15.258}      &\scalebox{0.98}{15.115} &\scalebox{0.98}{15.012} &\scalebox{0.98}{\textbf{14.766}} &\scalebox{0.98}{18.234}   \\
&  \scalebox{0.98}{SMAPE(\%)} &\scalebox{0.98}{\textbf{25.052}}  &  \scalebox{0.98}{25.750 } &  \scalebox{0.98}{25.455}          &\scalebox{0.98}{25.899} &\scalebox{0.98}{26.482} &\scalebox{0.98}{25.801} &\scalebox{0.98}{31.571}   \\

    \bottomrule
  \end{tabular}
    \end{small}
  \end{threeparttable}
\end{table}

\begin{table*}
    \centering
    \caption{Results of Cross-Dataset Transfer  Forecasting.}
    \label{tab:cross_transfer}
    \begin{minipage}{0.49\textwidth}
        \centering
  \begin{threeparttable}
  \begin{small}
  \renewcommand{\multirowsetup}{\centering}
  \setlength{\tabcolsep}{1.5pt}
  
  \begin{tabular}{c|c|ccccccc}
    \toprule
\multicolumn{2}{c|}{Source: PaaS} &  
{\rotatebox{0}{\scalebox{1.0}{{Fremer}}}} &
{\rotatebox{0}{\scalebox{1.0}{{PDF}}}} &
{\rotatebox{0}{\scalebox{1.0}{{iTrans.}}}} &
{\rotatebox{0}{\scalebox{1.0}{{PatchTST}}}} &
{\rotatebox{0}{\scalebox{1.0}{{FITS}}}} &
{\rotatebox{0}{\scalebox{1.0}{{DLinear}}}} &
{\rotatebox{0}{\scalebox{1.0}{{NLinear}}}}  \\
\midrule

\multirow{2}{*}{{\scalebox{0.95}{FaaS}}} 
&  \scalebox{0.98}{MSE} &\scalebox{0.98}{\textbf{0.308}}   &  \scalebox{0.98}{ 0.361 }  & \scalebox{0.98}{0.377}  &\scalebox{0.98}{0.312} &\scalebox{0.98}{0.322} &\scalebox{0.98}{0.312} &\scalebox{0.98}{0.333}   \\
&  \scalebox{0.98}{SMAPE(\%)} &\scalebox{0.98}{\textbf{9.204}} &   \scalebox{0.98}{9.932 } &  \scalebox{0.98}{11.233}  &\scalebox{0.98}{9.383} &\scalebox{0.98}{9.780} &\scalebox{0.98}{9.436} &\scalebox{0.98}{10.014}   \\
\midrule
\multirow{2}{*}{{\scalebox{0.95}{IaaS}}} 
&  \scalebox{0.98}{MSE} &\scalebox{0.98}{\textbf{0.763}}   &  \scalebox{0.98}{ 0.810} &\scalebox{0.98}{1.027} &\scalebox{0.98}{0.798} &\scalebox{0.98}{0.780} &\scalebox{0.98}{0.764} &\scalebox{0.98}{0.966}   \\
&  \scalebox{0.98}{SMAPE(\%)} &\scalebox{0.98}{\textbf{17.360}}   & \scalebox{0.98}{18.254 }  & \scalebox{0.98}{21.469} &\scalebox{0.98}{18.021} &\scalebox{0.98}{17.649} &\scalebox{0.98}{17.478} &\scalebox{0.98}{19.781}   \\
\midrule
\multirow{2}{*}{{\scalebox{0.95}{RDS}}} 
&  \scalebox{0.98}{MSE} &\scalebox{0.98}{\textbf{1.539}}   & \scalebox{0.98}{1.643 }  & \scalebox{0.98}{1.898} &\scalebox{0.98}{1.566} &\scalebox{0.98}{1.856} &\scalebox{0.98}{1.607} &\scalebox{0.98}{1.516}   \\
&  \scalebox{0.98}{SMAPE(\%)} &\scalebox{0.98}{\textbf{9.634}}   &  \scalebox{0.98}{ 10.539} & \scalebox{0.98}{13.474} &\scalebox{0.98}{10.007} &\scalebox{0.98}{10.268} &\scalebox{0.98}{9.965} &\scalebox{0.98}{10.854}   \\

    \bottomrule
  \end{tabular}
    \end{small}
  \end{threeparttable}
    \end{minipage}
    \hfill
    \begin{minipage}{0.49\textwidth}
        \centering
         \begin{threeparttable}
  \begin{small}
  \renewcommand{\multirowsetup}{\centering}
  \setlength{\tabcolsep}{1.5pt}
  
  \begin{tabular}{c|c|ccccccc}
    \toprule
\multicolumn{2}{c|}{Source: RDS} &  
             {\rotatebox{0}{\scalebox{1.0}{{Fremer}}}} &
{\rotatebox{0}{\scalebox{1.0}{{PDF}}}} &
{\rotatebox{0}{\scalebox{1.0}{{iTrans.}}}} &
{\rotatebox{0}{\scalebox{1.0}{{PatchTST}}}} &
{\rotatebox{0}{\scalebox{1.0}{{FITS}}}} &
{\rotatebox{0}{\scalebox{1.0}{{DLinear}}}} &
{\rotatebox{0}{\scalebox{1.0}{{NLinear}}}}  \\
\midrule

\multirow{2}{*}{{\scalebox{0.95}{FaaS}}} 
&  \scalebox{0.98}{MSE} &\scalebox{0.98}{\textbf{0.311}}&     \scalebox{0.98}{ 0.317 }   &    \scalebox{0.98}{0.313}  &\scalebox{0.98}{0.328} &\scalebox{0.98}{0.323} &\scalebox{0.98}{0.316} &\scalebox{0.98}{0.326}   \\
&  \scalebox{0.98}{SMAPE(\%)} &\scalebox{0.98}{\textbf{9.095}}&   \scalebox{0.98}{ 9.190 }    &  \scalebox{0.98}{ 9.704 }     &\scalebox{0.98}{9.507} &\scalebox{0.98}{10.151} &\scalebox{0.98}{9.405} &\scalebox{0.98}{9.475}   \\
\midrule
\multirow{2}{*}{{\scalebox{0.95}{IaaS}}} 
&  \scalebox{0.98}{MSE} &\scalebox{0.98}{\textbf{0.729}}&  \scalebox{0.98}{0.736  }    &   \scalebox{0.98}{ 0.891 }    &\scalebox{0.98}{0.788} &\scalebox{0.98}{0.760} &\scalebox{0.98}{0.756} &\scalebox{0.98}{0.810}   \\
&  \scalebox{0.98}{SMAPE(\%)} &\scalebox{0.98}{\textbf{16.696}}&  \scalebox{0.98}{16.751  }    &   \scalebox{0.98}{ 19.935 }    &\scalebox{0.98}{17.838} &\scalebox{0.98}{17.294} &\scalebox{0.98}{16.958} &\scalebox{0.98}{17.730}   \\
\midrule
\multirow{2}{*}{{\scalebox{0.95}{PaaS}}} 
&  \scalebox{0.98}{MSE} &\scalebox{0.98}{\textbf{0.063}}&    \scalebox{0.98}{ 0.064 }  &   \scalebox{0.98}{ 0.111 }    &\scalebox{0.98}{0.068} &\scalebox{0.98}{0.085} &\scalebox{0.98}{0.071} &\scalebox{0.98}{0.066}   \\
&  \scalebox{0.98}{SMAPE(\%)} &\scalebox{0.98}{\textbf{5.628}}&  \scalebox{0.98}{ 5.648 }   &   \scalebox{0.98}{ 12.148 }    &\scalebox{0.98}{6.461} &\scalebox{0.98}{8.929} &\scalebox{0.98}{6.606} &\scalebox{0.98}{6.222}   \\

    \bottomrule
  \end{tabular}
    \end{small}
  \end{threeparttable}
    \end{minipage}
\end{table*}



\subsection{RQ3: Generalizability Test}
To comprehensively assess the generalizability of \model, under the two experimental setups, the intra- and cross-dataset transfer, we validate the performance of \model. Note that models requiring fixed channels are unsuitable for transfer learning, as the number of channels in training and testing datasets often differs.

\begin{table}[t]
  \caption{{Results of Efficiency Test on Transformer-based Models."Training" means training time (ms) per iteration, "Inference" means inference time (ms) per iteration, and "Parameters" means parameter count (M).} }
  \label{tab:effi}
  \centering
  \begin{threeparttable}
  \begin{small}
  \renewcommand{\multirowsetup}{\centering}
  \setlength{\tabcolsep}{1.0pt}
  
  \begin{tabular}{c|c|cccccccc}
    \toprule
\multicolumn{2}{c|}{Models} &  
{\rotatebox{0}{\scalebox{1.0}{{Fremer}}}} &
{\rotatebox{0}{\scalebox{1.0}{{Fred.}}}} &
{\rotatebox{0}{\scalebox{1.0}{{PDF}}}} &
{\rotatebox{0}{\scalebox{1.0}{{iTrans.}}}} &
{\rotatebox{0}{\scalebox{1.0}{{PatchTST}}}} &
{\rotatebox{0}{\scalebox{1.0}{{Cross.}}}} &
{\rotatebox{0}{\scalebox{1.0}{{FED.}}}} &
{\rotatebox{0}{\scalebox{1.0}{{In.}}}} \\
\midrule

\multirow{4}{*}{{\scalebox{0.95}{IaaS}}} &
\scalebox{0.98}{Training} &\scalebox{0.98}{{61.67}} & \scalebox{0.98}{214.56} & \scalebox{0.98}{126.01}  &  \scalebox{0.98}{91.92} &\scalebox{0.98}{346.47} &\scalebox{0.98}{236.83} &\scalebox{0.98}{437.42} &\scalebox{0.98}{82.48}    \\
 & \scalebox{0.98}{Inference} 
&\scalebox{0.98}{{34.39}}& \scalebox{0.98}{92.44} & \scalebox{0.98}{42.97} &  \scalebox{0.98}{36.87}  &\scalebox{0.98}{47.77} &\scalebox{0.98}{91.56} &\scalebox{0.98}{214.45} &\scalebox{0.98}{62.39}   \\
& \scalebox{0.98}{Parameter} &\scalebox{0.98}{{0.57}}& \scalebox{0.98}{{111.13}} &\scalebox{0.98}{6.04}   &  \scalebox{0.98}{0.76} &\scalebox{0.98}{6.85} &\scalebox{0.98}{11.99} &\scalebox{0.98}{14.12} &\scalebox{0.98}{0.60}    \\
& \scalebox{0.98}{SMAPE(\%)} &\scalebox{0.98}{{16.25}} & \scalebox{0.98}{16.93} & \scalebox{0.98}{17.03}  & \scalebox{0.98}{16.71}   &\scalebox{0.98}{17.08} &\scalebox{0.98}{22.52} &\scalebox{0.98}{21.74} &\scalebox{0.98}{19.99}   \\
\midrule

\multirow{4}{*}{{\scalebox{0.95}{RDS}}} & \scalebox{0.98}{Training} &\scalebox{0.98}{{105.23}}& \scalebox{0.98}{5831.20} & \scalebox{0.98}{601.12}   &  \scalebox{0.98}{270.03} &\scalebox{0.98}{560.41} &\scalebox{0.98}{427.93} &\scalebox{0.98}{424.05} &\scalebox{0.98}{85.36}    \\
& \scalebox{0.98}{Inference} &\scalebox{0.98}{{27.87}}& \scalebox{0.98}{3006.61} & \scalebox{0.98}{235.19}  & \scalebox{0.98}{103.29}  &\scalebox{0.98}{32.09} &\scalebox{0.98}{100.09} &\scalebox{0.98}{186.10} &\scalebox{0.98}{50.46}    \\
& \scalebox{0.98}{Parameter} &\scalebox{0.98}{{0.59}}& \scalebox{0.98}{113.46} & \scalebox{0.98}{7.77}  & \scalebox{0.98}{0.77}  &\scalebox{0.98}{4.38} &\scalebox{0.98}{5.93} &\scalebox{0.98}{2.28} &\scalebox{0.98}{4.04}   \\
& \scalebox{0.98}{SMAPE(\%)} &\scalebox{0.98}{{8.66}}& \scalebox{0.98}{9.86} & \scalebox{0.98}{9.70}   & \scalebox{0.98}{9.15}   &\scalebox{0.98}{10.71} &\scalebox{0.98}{25.21} &\scalebox{0.98}{27.25} &\scalebox{0.98}{20.97}   \\

    \bottomrule
  \end{tabular}
    \end{small}
  \end{threeparttable}
\end{table}

\subsubsection{Intra-Dataset Transfer.} In this experimental setup, the models are trained using the training set of specific instances within a dataset. Subsequently, their performance is evaluated using the test set of the remaining instances. By default, the ratio of the number of training instances to test instances is set at 8:2. As can be seen from the results presented in Table \ref{tab:in_transfer}, \model\ achieves the best performance. This remarkable result clearly demonstrates the strong generalizability of \model\ when making predictions on workload series from unseen instances within the same dataset. \model\ can effectively capture the underlying patterns and characteristics present in the training data and apply them to new, unencountered data segments within the same dataset, showcasing its adaptability and robustness.

\subsubsection{Cross-Dataset Transfer.} 
In this experimental setup, models are trained on the training set of all instances from one dataset and evaluated on the test set of all instances from another dataset. Detailed results are presented in Table \ref{tab:cross_transfer}. 
Specifically, models are trained on PaaS and RDS datasets separately and then evaluated on other datasets.
Table \ref{tab:cross_transfer} demonstrates that \model\ consistently achieves the best performance across all experimental settings. This highlights \model’s exceptional zero-shot forecasting ability and underscores its potential as a foundational backbone for large-scale workload forecasting models. \model’s ability to generalize across datasets, capturing both commonalities and unique features of workload data, is a critical characteristic for building more comprehensive and powerful forecasting models.

\subsection{RQ4: Efficiency-Effectiveness Analysis}\label{sec:efficiency_test}

In this section, we evaluate the efficiency of \model\ in workload forecasting experiments, comparing it directly with other Transformer-based models. For a thorough analysis, we chose two datasets of different sizes: the smaller IaaS dataset with 93 instances, and the larger RDS dataset with 1113 instances. The efficiency evaluation results are summarized in Table \ref{tab:effi}.
From the table, it is clear that \model\ significantly outperforms other Transformer-based methods in terms of efficiency. Compared to the state-of-the-art model PatchTST, \model\ achieves notable improvements. For the IaaS dataset, it reduces training time by 82.2\%, inference time by 28.0\%, and parameter size by 91.7\%. Similarly, for the RDS dataset, it cuts training time by 81.2\%, inference time by 13.1\%, and parameter size by 86.5\%.
{Frequency domain methods (e.g., FredFormer, PDF, FEDformer) are often less efficient than time domain methods due to FFT-related overhead. However, \model, with its efficient design, achieves the best balance between efficiency and effectiveness.}
These improvements are attributed to \model’s lightweight components: the Complex-valued Spectrum Attention (CSA) mechanism and the frequency filter. The CSA efficiently extracts relevant information from the frequency domain while maintaining computational simplicity. The frequency filter optimizes data flow by selectively processing frequency components, minimizing unnecessary computational overhead.
Moreover, the results underscore \model’s advantage: superior forecasting performance and exceptional efficiency. This combination is crucial for real-world applications, where accurate and efficient workload forecasting is essential. \model\ thus emerges as a robust solution for addressing challenges across diverse industrial settings.

\begin{table*}[t]
\caption{Multivariate Forecasting Results on Datasets of Various Domains. }
\label{tab:general}

\begin{footnotesize}
\setlength{\tabcolsep}{1pt}
\begin{tabular}{cccc|cc|cc|cc|cc|cc|cc|cc|cc|cc|cc|cc|cc}
\toprule
\multicolumn{4}{c}{\scalebox{1.0}{Models}} &
\multicolumn{2}{c}{\rotatebox{0}{\scalebox{1.0}{{\model}}}} &
\multicolumn{2}{c}{\rotatebox{0}{\scalebox{1.0}{{Fredformer}}}} &
\multicolumn{2}{c}{\rotatebox{0}{\scalebox{1.0}{{PDF}}}} &
\multicolumn{2}{c}{\rotatebox{0}{\scalebox{1.0}{{iTransformer}}}} &
\multicolumn{2}{c}{\rotatebox{0}{\scalebox{1.0}{{PatchTST}}}} &
\multicolumn{2}{c}{\rotatebox{0}{\scalebox{1.0}{{Crossformer}}}} &
\multicolumn{2}{c}{\rotatebox{0}{\scalebox{1.0}{{FEDformer}}}} &
\multicolumn{2}{c}{\rotatebox{0}{\scalebox{1.0}{{Informer}}}} &
\multicolumn{2}{c}{\rotatebox{0}{\scalebox{1.0}{{DLinear}}}} &
\multicolumn{2}{c}{\rotatebox{0}{\scalebox{1.0}{{NLinear}}}} &
\multicolumn{2}{c}{\rotatebox{0}{\scalebox{1.0}{{MICN}}}} &
\multicolumn{2}{c}{\rotatebox{0}{\scalebox{1.0}{{FECAM}}}} \\
\toprule

& \multicolumn{1}{c@{\hspace{3pt}}@{\hspace{3pt}}}{Dataset} & Season. & Corr. & MAE & MSE& MAE & MSE & MAE & MSE& MAE & MSE & MAE & MSE & MAE & MSE & MAE & MSE & MAE & MSE & MAE & MSE & MAE & MSE & MAE & MSE & MAE & MSE  \\
\midrule

\multirow[c]{5}{*}{{\rotatebox{90}{\scalebox{1.0}{{Strong Periodicity}}}}}& \multirow[c]{1}{*}{{Electricity}}& 0.945 & 0.802 &\textbf{0.256} & \textbf{0.161} & 0.279 & 0.181 & \underline{0.260} & 0.164 & 0.270 & 0.194 & {0.261} & \underline{0.163} & 0.279 & 0.181 & 0.324 & 0.211 & 0.364 & 0.265 & 0.264 & 0.167 & 0.261 & 0.169 & 0.288 & 0.179 & 0.332 & 0.238 \\
\cmidrule(lr){2-28}

& \multirow[c]{1}{*}{{Solar}}& 0.919 & 0.753  & \textbf{0.257} & \textbf{0.195}  & 0.287 & 0.226 & 0.264 & \underline{0.200} &\underline{0.262} & {0.202 } & 0.294 & {0.207} & 0.365 & 0.330 & 0.482 & 0.421 & 0.397 & 0.380 & 0.309 & 0.247 & {0.272} & 0.255 & 0.296 & 0.248 & 0.315 & 0.261 \\
\cmidrule(lr){2-28}

& \multirow[c]{1}{*}{{Traffic}}& 0.880 & 0.813 &\textbf{0.273} & \textbf{0.389} & 0.339 & 0.517 & \underline{0.279} & 0.399 & {0.281} & \underline{0.397}& {0.283} & {0.405} & 0.284 & 0.523 & 0.377 & 0.615 & 0.414 & 0.752 & 0.295 & 0.434 & 0.290 & 0.433 & 0.308 & 0.534 & 0.445 & 0.715 \\
\cmidrule(lr){2-28}


& \multirow[c]{1}{*}{{PEMS04}}& 0.854 & 0.797  & \textbf{0.241} & \textbf{0.142}  & 0.433 & 0.361 & 0.305 & 0.205 & \underline{0.261}&  \underline{0.156}& 0.303 & 0.191 & {0.269} & {0.168} & 0.709 & 0.867 & 0.338 & 0.236 & 0.332 & 0.243 & 0.330 & 0.255 & 0.445 & 0.388 & 0.443 & 0.369 \\
\cmidrule(lr){2-28}

& \multirow[c]{1}{*}{{PEMS08}}& 0.850 & 0.807 & \textbf{0.265} & 0.288   & 0.486 & 0.582 & 0.326 & 0.373 &\underline{0.271} &0.298 & 0.306 & \underline{0.213} & {0.273} & \textbf{0.176} & 0.706 & 0.876 & 0.402 & 0.350 & 0.363 & 0.298 & 0.360 & 0.322 & 0.472 & 0.436 & 0.493 & 0.620\\
\midrule


\multirow[c]{6}{*}{{\rotatebox{90}{\scalebox{1.0}{{Weak Periodicity\quad}}}}}& \multirow[c]{1}{*}{{ZafNoo}}& 0.757 & 0.598 & 0.464 & 0.522   & 0.467 & 0.595 & 0.453 & 0.515 &0.456& 0.523  & 0.465 & 0.511 & 0.455 & \textbf{0.494} & 0.499 & 0.578 & 0.602 & 0.744 & \textbf{0.451} & 0.496 & 0.458 & 0.522 & \underline{0.452} & \underline{0.495} & 0.520 & 0.717\\
\cmidrule(lr){2-28}


& \multirow[c]{1}{*}{{ETTh1}}& 0.730 & 0.630  & 0.446 & 0.435 &  0.438  & 0.449 & \underline{0.426} & \textbf{0.407} &  0.448& 0.439 & \underline{0.428} & {0.411} & 0.467 & 0.453 & 0.454 & 0.432 & 0.583 & 0.731 & 0.432 & 0.419 & \textbf{0.421} & \underline{0.410} & 0.449 & 0.423 & 0.560 & 0.674\\
\cmidrule(lr){2-28}

& \multirow[c]{1}{*}{{AQShunyi}}& 0.720 & 0.612  & 0.522 & 0.723  & 0.531 & 0.819 & 0.507 & \underline{0.703} &\textbf{0.503} & 0.706 & {0.509} & \underline{0.705} & \underline{0.504} & \textbf{0.694} & 0.546 & 0.763 & 0.545 & 0.782 & 0.522 & 0.706 & 0.514 & 0.713 & 0.534 & 0.735  & 0.549 & 0.775  \\
\cmidrule(lr){2-28}

& \multirow[c]{1}{*}{{Weather}}& 0.652 & 0.663  & 0.284 & 0.241 & 0.272 & 0.244 & \underline{0.263} & \underline{0.227} & {0.270} & {0.232}  & \textbf{0.262} & \textbf{0.225} & 0.294 & {0.235} & 0.351 & 0.306 & 0.323 & 0.300 & 0.289 & 0.239 & {0.281} & 0.249 & 0.288 & 0.238 & 0.305 & 0.260 \\
\cmidrule(lr){2-28}



& \multirow[c]{1}{*}{{PEMS-BAY}}& 0.618 & 0.842  & \underline{0.375} & \textbf{0.577}  & 0.454 & 0.759 & 0.399 & 0.688 &0.381 & \underline{0.583}  & 0.398 & 0.663 & \textbf{0.374} & {0.590} & 0.598 & 0.959 & 0.460 & 0.899 & 0.443 & 0.719 & 0.445 & 0.748 & 0.476 & 0.854 & 0.544 & 0.959\\
\cmidrule(lr){2-28}


& \multirow[c]{1}{*}{{METR-LA}}& 0.490  & 0.778  & 0.733 & 1.274 & 0.734 & 1.484    &0.728 & 1.275 & 0.720 & 1.354 & \underline{0.704} & \underline{1.250} & \textbf{0.695} & 1.357 & 0.865 & 1.671 & 0.724 & 1.606 & 0.727 & \textbf{1.203} & 0.751 & 1.314 & 0.727 & 1.373 & 0.753 & 1.288 \\



\bottomrule
\end{tabular}
\end{footnotesize}
\end{table*}
\subsection{RQ5: Extended to General Forecasting}
Although \model\ was initially designed for workload forecasting, our research demonstrates its effectiveness across diverse domains. To explore this further, we select widely-used datasets from TFB \citep{qiu2024tfb}, spanning domains such as traffic flow prediction and electricity consumption forecasting, and evaluate \model’s performance on these datasets.
Table \ref{tab:general} presents the average forecasting results across four horizons $\{96, 192, 336, 720\}$, with all baseline results directly referenced from the TFB paper \citep{qiu2024tfb}. It could be observed that \model\ outperforms all baseline models on datasets with strong periodicity, such as Traffic, Electricity, and PEMS04. This exceptional performance underscores \model’s potential in forecasting tasks characterized by periodicity, leveraging its frequency-domain design to effectively extract global dependencies in series. These findings suggest that \model can capture generalizable patterns beyond workload series, making it a versatile tool for applications ranging from traffic management to energy planning.

\subsection{RQ6: Predictive Auto-Scaling}

In this section, we perform proactive (predictive) Horizontal Pod Autoscaler (HPA) scaling simulation tests on a Kubernetes cluster.
For the forecasting model, a 5-day historical window is employed to predict the next-day workload. Based on this predicted value, a 24-hour simulation test is carried out to assess the quality of service and resource overhead when applying different forecasting models to Kubernetes HPA's proactive auto-scaling.
The results of the simulation experiment are presented in Table \ref{tab:hpa}. These include delays at different quantiles (for evaluating service quality) and the average and maximum number of Pods (for evaluating resource consumption). Here, Naïve HPA denotes the native passive (responsive) HPA in Kubernetes. Ideal represents the HPA that utilizes real values instead of predicted values for proactive (predictive) scaling, and it represents, to some extent, the performance upper-bound of predictive scaling. Ave-Lat represents the average latency, x-Lat represents the x-quantile latency, both measured in seconds (s). Timeout Rate denotes the number of requests that exceeds the timeout threshold, which is set to 10s in this experiment. AvePod represents the average number of Pods consumed.
We utilize the workload series associated with a function instance in the ByteDance FaaS service for testing. Its workload data is also part of the FaaS dataset used to evaluate the workload-forecasting performance in section \ref{exp:forecasting}. Specifically, we record the workload and replay it from a client. The scaling strategy adheres to the original HPA mechanism, and is based on the assumption that the workload volume has a linear relationship with resource consumption. 

\begin{table}[ht]
  \caption{Kubernetes HPA Test Results}
  \label{tab:hpa}
  \centering
  \begin{threeparttable}
  \renewcommand{\multirowsetup}{\centering}
  \setlength{\tabcolsep}{1pt}
  \begin{tabular}{c|c|c|c|c|c|c}
    \toprule
    \scalebox{0.78}{Models} & \scalebox{0.78}{Ave-Lat(s)}  & \scalebox{0.78}{99.9-Lat(s)} & \scalebox{0.78}{99-Lat(s)} & \scalebox{0.78}{90-Lat(s)} & \scalebox{0.78}{Timeout Rate} & \scalebox{0.78}{AvePod} \\
    \midrule
    \scalebox{0.78}{PatchTST} & \scalebox{0.78}{1.017} & \scalebox{0.78}{10.0} & \scalebox{0.78}{3.644} & \scalebox{0.78}{1.651} & \scalebox{0.78}{0.132\%} & \scalebox{0.78}{22.479} \\
    \scalebox{0.78}{DLinear} & \scalebox{0.78}{1.081} & \scalebox{0.78}{10.0} & \scalebox{0.78}{4.108} & \scalebox{0.78}{1.779} & \scalebox{0.78}{0.22\%} & \scalebox{0.78}{19.25} \\
    \scalebox{0.78}{FITS} & \scalebox{0.78}{1.05} & \scalebox{0.78}{10.0} & \scalebox{0.78}{4.101} & \scalebox{0.78}{1.704} & \scalebox{0.78}{0.279\%} & \scalebox{0.78}{21.889} \\
    \midrule
    \scalebox{0.78}{Naïve HPA} & \scalebox{0.78}{0.996} & \scalebox{0.78}{10.0} & \scalebox{0.78}{3.764} & \scalebox{0.78}{1.56} & \scalebox{0.78}{0.386\%} & \scalebox{0.78}{29.181} \\
    \midrule
    \scalebox{0.78}{Ideal} & \scalebox{0.78}{0.789} & \scalebox{0.78}{3.513} & \scalebox{0.78}{2.063} & \scalebox{0.78}{1.206} & \scalebox{0.78}{0.026\%} & \scalebox{0.78}{21.382} \\
    \midrule
    \scalebox{0.78}{Fremer} & \scalebox{0.78}{0.826} & \scalebox{0.78}{10.0} & \scalebox{0.78}{2.292} & \scalebox{0.78}{1.261} & \scalebox{0.78}{0.102\%} & \scalebox{0.78}{21.951} \\
    \bottomrule
\end{tabular}
  \end{threeparttable}
  
\end{table}

Through a comprehensive examination and analysis of the experimental results, we conclude that the Fremer model demonstrates exceptional efficacy in forecasting future workloads. It accurately identifies trends in workload variations, thereby providing robust support for enhancing Quality of Service (QoS). This improvement is particularly evident in delay metrics across diverse quantiles. The Fremer model significantly outperforms other proactive auto-scaling strategies based on alternative forecasting models. In practical applications, this implies that employing the Fremer model for workload prediction can more effectively ensure the smooth operation and stability of services, reduce user waiting times, and enhance the overall user experience.

Additionally, Fremer achieves remarkable results in optimizing resource utilization, as clearly reflected in the fluctuations of the average Pod count. For instance, comparative data analysis reveals that the Fremer model reduces average latency by 18.78\% compared to the PatchTST model, while utilizing 2.35\% fewer Pods on average. This indicates that the Fremer model can maintain service quality while using computing resources more efficiently, minimizing resource waste, and reducing operational costs for enterprises.

However, we also note that proactive Horizontal Pod Autoscalers (HPAs) guided by other forecasting models exhibit strong performance in specific metrics. For example, the proactive HPA strategy based on the DLinear model achieves the lowest average number of Pods utilized. Nevertheless, its average latency is the highest among the compared models. Through detailed analysis, we attribute this to the DLinear’s forecasts being significantly lower than the actual workload during certain periods. Consequently, the replica number recommendations derived from its predictions are insufficient to handle the actual workload, resulting in increased service latency.


\section{Conclusion} 

{We propose \model, an efficient and effective deep forecasting model addressing critical challenges in cloud service workload forecasting. By utilizing frequency domain representations, \model balances accuracy, efficiency, and generalizability, meeting modern cloud requirements. Its innovations—Learnable Linear Padding, Frequency Filters, and Complex-valued Spectrum Attention—enable superior performance over state-of-the-art models with reduced computational costs.
We also release four large-scale, high-quality datasets collected from ByteDance’s cloud services, covering thousands of computing instances over 1–2 months, providing robust resources for research and benchmarking.
\model addresses Transformer limitations and demonstrates strong generalizability, advancing cloud service forecasting. Future work will focus on scalable models for large cloud systems.}


\clearpage

\bibliographystyle{ACM-Reference-Format}
\bibliography{sample}

\end{document}